\documentclass[sigconf]{aamas} 

\usepackage{balance} 
\usepackage{hyperref}
\usepackage{url}
\usepackage{booktabs}       
\usepackage{amsfonts}       
\usepackage{nicefrac}       
\usepackage{microtype}      
\usepackage{xcolor}         
\usepackage{amsmath, amsthm}
\usepackage{graphicx}
\usepackage{subfigure}
\usepackage{enumitem}
\usepackage{wrapfig}
\usepackage{algorithmicx, algorithm, algpseudocode}

\newtheorem{theorem}{Theorem.}
\newtheorem{corollary}{Corollary.}
\newtheorem{lemma}{Lemma.}

\title{Dual Alignment Maximin Optimization for Offline MBRL}

\author{Chi Zhou}
\affiliation{
  \institution{UCAS}
  \country{China}}
\email{Zhouchi23@mails.ucas.ac.cn}

\author{Wang Luo}
\affiliation{
  \institution{UCAS}
  \country{China}}

\author{Haoran Li}
\affiliation{
  \institution{UCAS}
  \country{China}}

\author{Congying Han}
\affiliation{
  \institution{UCAS}
  \country{China}}

\author{Tiande Guo}
\affiliation{
  \institution{UCAS}
  \country{China}}

\author{Zicheng Zhang}
\affiliation{
  \institution{JD}
  \country{China}}

\begin{abstract}
    Offline reinforcement learning agents face significant deployment challenges due to the synthetic-to-real distribution mismatch.
    While most prior research has focused on improving the fidelity of synthetic sampling and incorporating off-policy mechanisms, the directly integrated paradigm often fails to ensure consistent policy behavior in biased models and underlying environmental dynamics, which arise from discrepancies between behavior and learning policies.
    In this paper, we first shift the focus from model reliability to constraining policy discrepancies while optimizing for expected returns, and then self-consistently incorporate synthetic data, deriving a novel maximin paradigm, \textbf{D}ual \textbf{A}lignment \textbf{M}aximin \textbf{O}ptimization~(DAMO).
    DAMO is a unified framework to ensure both \textit{model-environment policy consistency} and \textit{synthetic and offline data compatibility}.
    The inner minimization performs dual conservative value estimation, aligning policies and trajectories to avoid out-of-distribution states and actions, while the outer maximization ensures that policy improvements remain consistent with inner value estimates.
    Empirical evaluations demonstrate that DAMO effectively ensures model and policy alignments, achieving competitive performance across diverse offline benchmark tasks.
\end{abstract}


\newcommand{\BibTeX}{\rm B\kern-.05em{\sc i\kern-.025em b}\kern-.08em\TeX}

\begin{document}


\pagestyle{fancy}
\fancyhead{}


\maketitle 


\section{Introduction}

    Offline reinforcement learning~(RL)~\cite{lange2012batch, levine2020offline} aims to learn policies from a pre-collected dataset generated by a behavior policy within the real environment. 
    This paradigm helps avoid the potential safety risks and high costs associated with direct online interactions, making RL applicable in real-life scenarios, such as healthcare decision-making support and autonomous driving~\cite{emerson2023offline, sinha2022s4rl, mnih2015human}. 
    Offline model-based RL~(MBRL)~\cite{mopo, morel, mobile} improves upon this by training a dynamics model and using it to generate synthetic data for policy training. 
    The introduction of the dynamics model enhances the sample efficiency and allows the agent to answer counterfactual queries~\cite{levine2020offline}. 
    However, despite strong performance in dynamics models, trained policies often degrade when deployed in real scenarios due to mismatches between synthetic and real distributions.

\begin{figure}[t]
        \centering
        \subfigure{
        \begin{minipage}[b]{\linewidth}
            \centering
            \includegraphics[width=\linewidth]{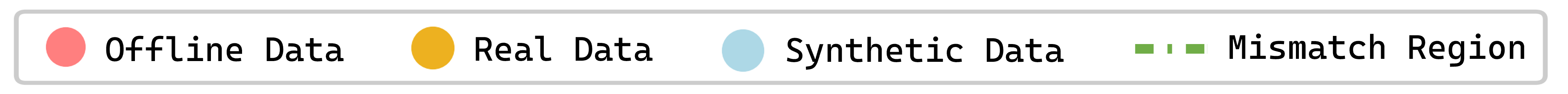}
        \end{minipage}
        }
        \setcounter{subfigure}{0}
        \subfigure[MOPO]{
        \begin{minipage}{0.3\linewidth}
            \centering
            \includegraphics[width=\linewidth]{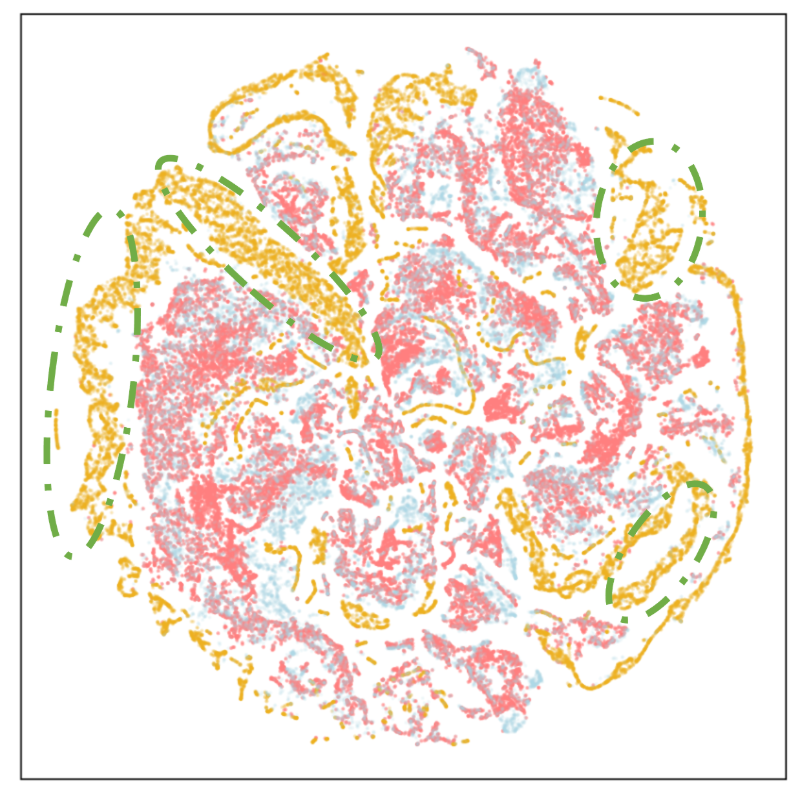}
        \end{minipage}
        }
        \subfigure[MOBILE]{
        \begin{minipage}{0.3\linewidth}
            \centering
            \includegraphics[width=\linewidth]{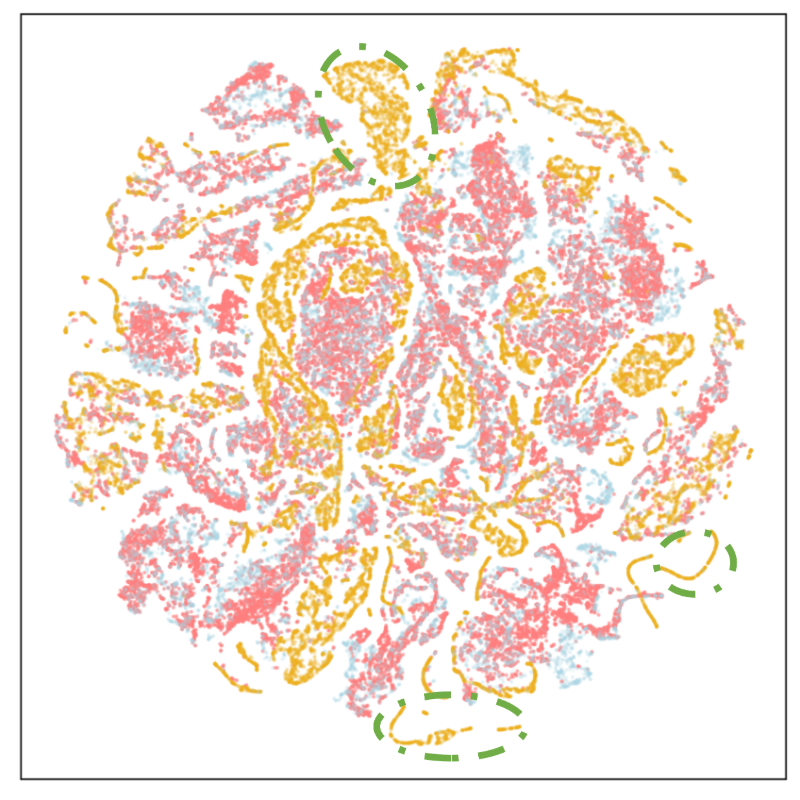}
        \end{minipage}
        }
        \subfigure[DAMO]{
        \begin{minipage}{0.3\linewidth}
            \centering
            \includegraphics[width=\linewidth]{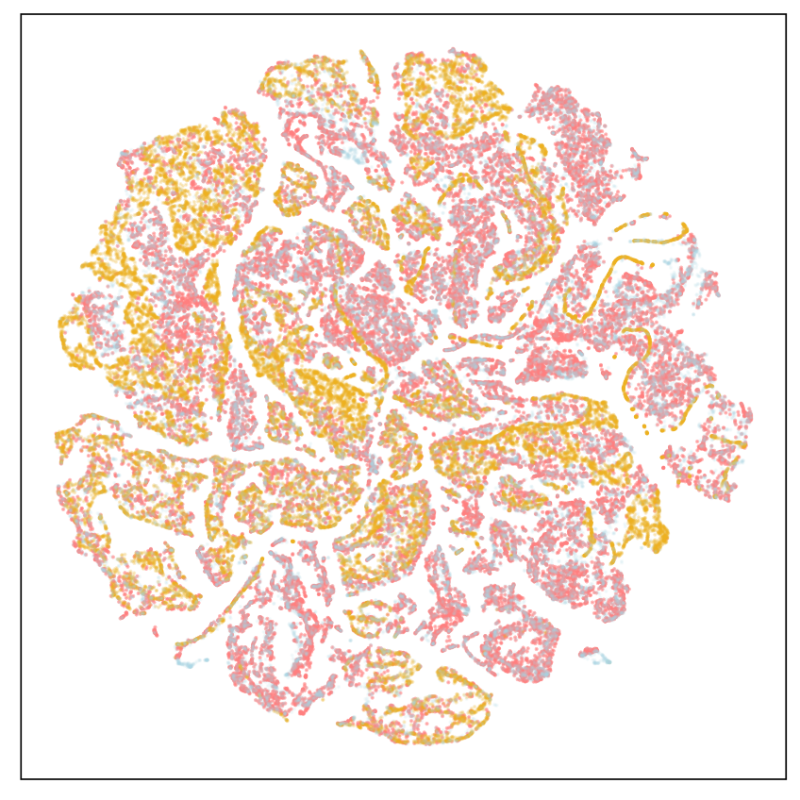}
        \end{minipage}
        }
        \caption{The distributions of transition pair $(s, a, s^{\prime})$ across three datasets (Offline Data, Real Data, and Synthetic Data) are shown for the hopper-medium-expert task, comparing MOPO and MOBILE with our proposed DAMO framework. 
        The Real Data is collected through executing the policy trained using specific offline methods in the real environment. 
        While MOPO and MOBILE align synthetic and offline data well (indicated by a high overlap), they fail to mitigate OOD states in the real environment (circled by an ellipse). 
        In contrast, DAMO effectively eliminates the OOD region.}
        
        \label{fig-mopo}
    \end{figure}

    Most prior works have focused on enhancing the reliability of synthetic samplings from dynamics models, and directly incorporating off-policy optimization methods such as SAC~\cite{haarnoja2018soft}, to address this mismatch. 
    The representative approach is MOPO~\cite{mopo}, which quantifies dynamics model uncertainty by measuring the predicted variance of learned models and penalizes synthetic trajectories with high uncertainty. 
    This methodology has inspired subsequent researchs like MOReL~\cite{morel} and MOBILE~\cite{mobile}, which adopt similar uncertainty-aware frameworks. 
    These approaches construct conservative value functions to mitigate overestimation of out-of-distribution~(OOD) actions, thereby reducing execution of these unreliable actions and facilitating the generation of synthetic data compatible with the offline dataset. 
    However, as illustrated in Fig.~\ref{fig-mopo}, although this uncertainty-aware framework aligns the synthetic data with offline data and mitigates OOD actions during policy training, it fails to alleviate OOD trajectories when deployed into real scenarios, leading to inconsistent policy behavior during deployment and leaving the OOD issues in offline MBRL unresolved.



    To bridge this gap, we start from a regularized policy optimization objective, which limits the discrepancy between the visitation distribution of the training policy and the behavior policy under the real environment. By self-consistently incorporating the synthetic data into the regularized objective, we derive a unified maximin optimization objective for offline MBRL. Building on this objective, we propose Dual Alignment Maximin Optimization~(DAMO), a unified actor-critic framework to ensure dual conservative estimation.

    DAMO iteratively applies maximin optimization in two phases: the inner minimization and the outer maximization. 
    In the inner minimization, equivalent to critic training, we provide a theoretical analysis showing that it enforces dual conservative estimation. This value estimation explicitly aligns the synthetic and offline data to prevent OOD actions during training, while implicitly aligning learning policy behaviors in both the learned model and the real environment to avoid OOD states during deployment.
    In the outer maximization, corresponding to actor learning, policy improvements are guided by the inner value estimates through objective alignment. 
    This preserves conservative behavior and mitigates OOD issues throughout both policy optimization and deployment.
    Additionally, DAMO incorporates a classifier to enable the dual conservative estimation during practical implementation.
    Extensive experiments showcase that DAMO successfully mitigates OOD issues and achieves competitive performance across various offline benchmarks. 
    Our contributions are summarized as follows:
    \begin{itemize}[topsep=1em]
        \item We derive a consistent maximin objective for offline MBRL, ensuring both alignments between \textit{dynamics model and real environment policy behaviors}, and \textit{synthetic and offline} data.
        \item We develop DAMO, an actor-critic framework that employs maximin policy optimization to achieve both the \textit{dual conservative value estimation} and the \textit{consistent policy improvement}.
        \item We conduct experiments to validate that DAMO alleviates the OOD issues throughout training and deployment, delivering superior performance across varying benchmarks.
    \end{itemize}

\section{Related Work}
    \textbf{Offline Model-based RL} trains a dynamics model using offline data to approximate environmental dynamics and performs conservative policy optimization~\cite{lu2021revisiting} with model-generate data to mitigate OOD issues. 
    MOPO~\cite{mopo} penalizes state-action pairs with high model uncertainty, effectively discouraging the agent from selecting OOD actions. 
    MOReL~\cite{morel} constructs a pessimistic Markov decision process to prevent the agent from entering OOD regions, ensuring safer and more reliable policy learning. 
    COMBO~\cite{combo} extends Conservative Q-learning~\cite{kumar2020conservative} to the model-based setting by regularizing the value function on OOD samples.
    RAMBO~\cite{rambo} employs an adversarial training framework, optimizing both the policy and model jointly to ensure accurate transition predictions while maintaining robustness. 
    MOBILE~\cite{mobile} incorporates penalties into value learning by utilizing uncertainty in Bellman function estimates derived from ensemble models.
    SAMBO~\cite{luo2024sambo} imposes reward penalties and fosters exploration by incorporating model and policy shifts inferred through a probabilistic inference framework.

    \quad

    \noindent
    \textbf{DICE-based Methods.} Distribution Correction Estimation~(DICE) is a technique that estimates stationary distribution ratios using duality theory.
    DICE has shown superior performance in evaluating discrepancies between distributions and has been widely applied in various RL domains, including off-policy evaluation~(OPE)~\cite{nachum2019dualdice}, offline imitation learning~(IL)~\cite{ma2022smodice}, and offline RL~\cite{algaedice}. 
    In offline RL, DICE-based methods correct distribution shifts between offline data and environmental dynamics, formulating a tractable maximin optimization objective. 
    For instance, AlgaeDICE~\cite{algaedice} pioneers the application of DICE-based methods in offline RL, employing a regularized dual objective and Lagrangian techniques to address distribution shift challenges. 
    OptiDICE~\cite{optidice} incorporates the Bellman flow constraint and directly estimates stationary distribution corrections for the optimal policy, eliminating the need for policy gradients. 
    ODICE~\cite{odice} deploys orthogonalization techniques to eliminate the conflicts between varying gradients, and updates the policy with DICE to enforce state-action-level constraints. 
    Unlike these proposed model-free approaches, DAMO is, to the best of our knowledge, the first method to promote DICE insights in the model-based setting. 
    While model-free methods typically handle discrepancies between two distributions, DAMO skillfully extends this concept by managing the discrepancies across three distributions, using a divergence upper bound to ensure dual alignment.

\section{Preliminaries}
    \textbf{MDP.}
        We consider the Markov decision process~(MDP) defined by the six-element tuple $M = (\mathcal{S}, \mathcal{A}, T, r, \mu_{0}, \gamma)$, where $\mathcal{S}$ denotes the state space, $\mathcal{A}$ denotes the action space, $T(s^{\prime} | s, a)$ is the environment transition, $r(s, a, s^{\prime}) > 0$ is the reward function, $\mu_{0}$ is the initial state distribution, and $\gamma \in (0, 1)$ is the discounted factor. Given an MDP, the objective of RL is to find a policy $\pi: \mathcal{S} \rightarrow \Delta(\mathcal{A})$ that maximizes the cumulative reward from the environment. This objective can be formally expressed as: $\pi^{\star} = \arg\max_{\pi} \mathbb{E}_{\mu_{0}, \pi, T } [\sum_{t=0}^{\infty} \gamma^{t} r(s_{t}, a_{t} ,s_{t+1}) ]$. In this paper, we utilize the dual form of the RL objective~\cite{puterman2014markov}, which is represented as follows:
    	\begin{equation*}
    		\pi^{\star} = \arg\max\limits_{\pi} \mathbb{E}_{ \rho^{\pi}_{T}} [r(s, a, s^{\prime})].
    	\end{equation*}
            
    	Here, $\rho^{\pi}_{T}(s, a, s^{\prime})$ is the transition occupancy measure, characterizing the probability distribution of transition pairs $(s, a, s^{\prime})$ induced by policy $\pi$ under the dynamics $T$ among the episode, defined by:
        \begin{multline*}
            \rho_{T}^{\pi} (s,a,s') = (1-\gamma) \sum_{t=0}^{\infty} \gamma^{t} \mathbb{P}[s_t = s , a_t = a , 
            s_{t+1} = s'
            \\
            | s_0 \sim \mu_{0} , 
            a_t \sim \pi(\cdot|s_t) , s_{t+1} \sim T(\cdot | s_t,a_t)].
        \end{multline*}
        
        $\mathbb{P}$ represents the probability. For simplicity, we will substitute $\rho^{\pi}_{T}(s, a, s')$ with notation $\rho^{\pi}_{T}$ in the remainder of the article.

\section{Demystify OOD in Offline MBRL}
        \begin{figure*}[t]
            \centering
            \includegraphics[width=0.9\linewidth]{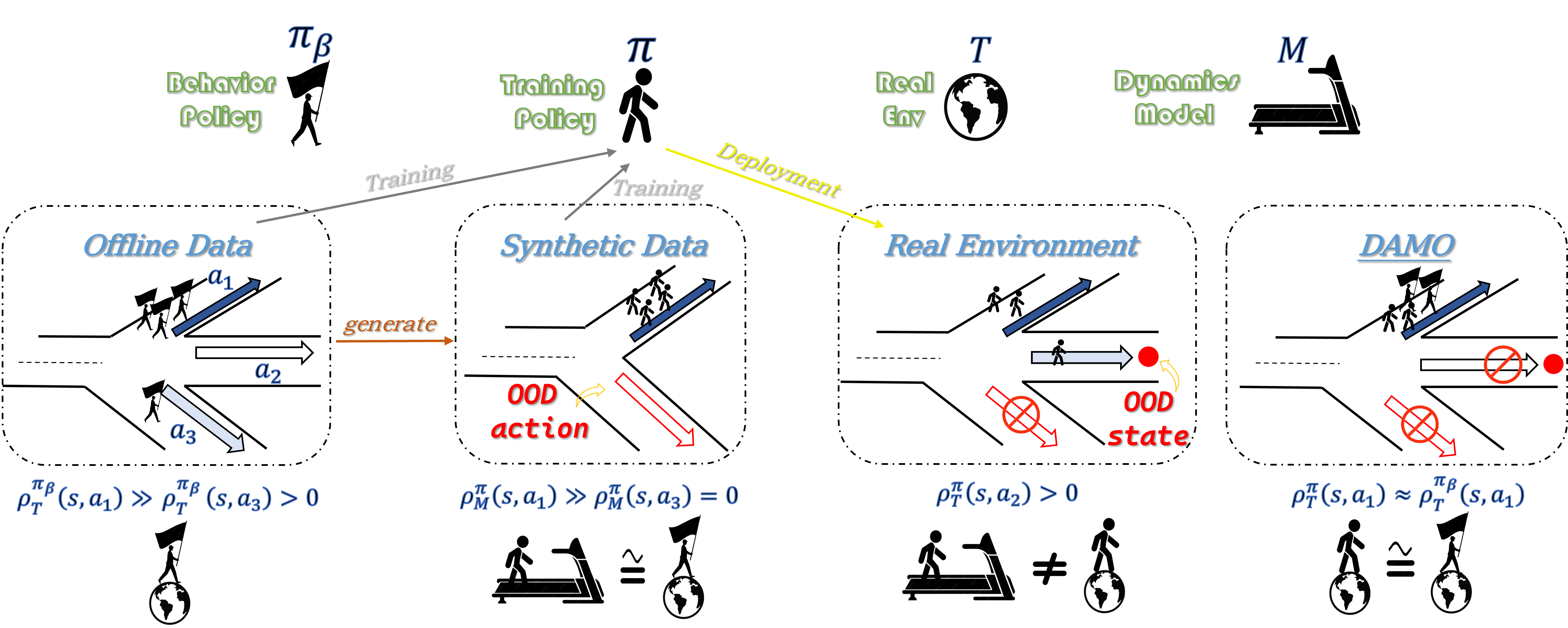}
            \caption{Intuitive examples for the OOD issue. 
            The behavior policy~$\pi_{\beta}$ primarily favors the top road~($a_{1}$) with a minor representation of the bottom road~($a_{3}$) and never selects the middle road~($a_{2}$). 
            Thus, the dynamics model lacks representation of the $a_{2}$ and treats the $a_{3}$ as an OOD action.
            The policy $\pi$, trained on both offline and synthetic data, avoids $a_{3}$ when deployed in the real environment but struggles to evaluate $a_{2}$ due to its absence in $M$, resulting in occasionally getting into OOD state. 
            In DAMO, $\pi$ is constrained by $\pi_{\beta}$, guiding the agent to consistently select $a_{1}$. At the bottom, we show that while data alignment ensures a similar distribution between offline and synthetic data, it fails to guarantee behavior consistency between $M$ and $T$. 
            }
            \label{fig-diagram}
        \end{figure*}
        
        In this section, as illustrated in Fig.~\ref{fig-diagram}, we explore the OOD issues in offline MBRL. We focus on the essence of aligning both the \textit{synthetic and offline data}, as well as the \textit{dynamics model and real environment policy behavior}, which motivates the direction of this research.

        \subsection{OOD Issues in Offline RL}
        In offline RL, OOD problems~\cite{mao2024offline} arise when agents encounter states or select actions that fall outside the offline data. Avoiding the OOD issue is a prerequisite for effective policy training.

        \quad

        \noindent
        \textbf{OOD actions} refer to actions that the behavior policy does not choose in specific states, rather than actions that are absent from the offline dataset.
        The value of OOD actions is prone to overestimation due to limited sample coverage of the offline dataset and the requirement of maximizing the action values, leading to the incorrect or suboptimal direction of policy improvement.  
        
        \quad

        \noindent
        \textbf{OOD states} primarily arise in three main scenarios:
        1) The policy executes unreliable OOD actions, leading to transitions into OOD states. 
        2) Stochastic dynamics could transition the agent into OOD states, even if it takes in-distribution~(ID) actions in ID states. Additionally, the initial state of the real environment may lie outside the dataset.
        3) Nonstationary dynamics, such as disturbances from real world~\cite{luo2024ompo} and cross-domain settings~\cite{lyu2025cross, lyu2024cross} can introduce fluctation in state distributions. 
        In this paper, we consider the first two factors.
        While the third factor is outside the scope of the offline RL setting, it is a critical consideration in some real-world applications.

    \subsection{OOD Issues under Biased Dynamics Model}
    \label{bsec-OOD actions and states in offline model-based RL}

        The prediction bias and stochasticity of the dynamics model compound the OOD issue in the offline setting. The prediction bias causes a misalignment between synthetic and offline data, and this bias is amplified with iterative use of the model, leading to the undesirable exploitation of OOD state-action pairs~\cite{levine2020offline}. Meanwhile, the stochasticity elevates the likelihood of encountering OOD states, undermining the policy performance across both learned models and real environments.
        More discussions are given in Appendix~\ref{app-bsec-OOD actions and states in offline model-based RL}.

        \quad
        
        \noindent
        \textbf{OOD actions.}
        The integration of the dynamics model allows the agent to query the possible outcomes of selecting arbitrary actions in any given state. However, the dynamics model cannot provide accurate predictions for OOD actions, and the prediction bias will mislead the estimation of these OOD action values. To address the problem, it is essential to constrain the scope of counterfactual queries within the support of the offline dataset, which amounts to aligning the distribution of synthetic data with the offline data.

        \quad
        
        \noindent
        \textbf{OOD states.}
        The agent can mitigate the OOD states stemming from the first scenario by constraining the training policy to avoid OOD actions during training. However, the inherent prediction bias and intrinsic stochasticity of the biased dynamics model will increase the potential of encountering OOD states stemming from the second scenario during deployment. To address this problem,  it is essential to ensure that the policy behavior remains consistent across both the dynamics model and the real environment.

        In summary, while optimizing the training policy $\pi$ to maximize the expected returns, we should constrain the policy in two directions: 1) aligning the synthetic data $(s, a, s^{\prime})_{M}^{\pi}$ with the offline data $(s, a, s^{\prime})_{T}^{\pi_{\beta}}$, and 2) ensuring that training policy $\pi$ exhibits consistent behaviors in both dynamics models~$M$ and real environments~$T$.

\section{\mbox{Dual Alignment Maximin Optimization}}
\label{sec-Method}
    Inspired by the analysis above, we propose Dual Alignment Maximin Optimization~(DAMO), a unified maximin approach to mitigate both OOD actions and states through dual alignment. 
    We further explore the distinct roles of the inner minimization and the outer maximization of DAMO in addressing the distribution shift.

    \subsection{Maximin Objective for Shift Mitigation}
    \label{bsec-A unified framework for offline model-based RL}
        
        We start from the root cause of distribution shifts, \textit{i.e.,} the discrepancy between learned and behavior policy behaviors in the underlying environment $T$.
        To tackle this, we imitate the behavior regularization methods and introduce a regularized objective that constrains the difference between distributions of learned and behavior policy while optimizing for expected rewards:
        \begin{equation}\notag
        	\max\limits_{\pi} \mathbb{E}_{\rho_{T}^{\pi}} [ r(s, a, s^{\prime}) ] - \alpha D_{KL}(\rho_{T}^{\pi} \Vert \rho_{T}^{\pi_{\beta}}),
        	\label{eq-align offline data with real data}
        \end{equation}
        where $\alpha$ is a regularization hyperparameter.
        Next, we self-consistently incorporate synthetic data through establishing an upper bound of KL divergence, deriving the following surrogate objective:
        \begin{equation}
        	\max\limits_{\pi} \mathbb{E}_{\rho_{T}^{\pi}} [ r(s, a, s^{\prime})-\alpha\log\dfrac{\rho_{M}^{\pi}}{\rho_{T}^{\pi_{\beta}}} 
        	]-\alpha D_{f}(\rho_{T}^{\pi} \Vert \rho_{M}^{\pi}).
        	\label{eq-surrogate objective}
        \end{equation}
        Where $f$ is a convex function that satisfies the inequality: $f(x) \geq x\log x, \forall x > 0.$ This surrogate objective integrates two essential elements: a data alignment regularizer $\log(\rho_{M}^{\pi}/\rho_{T}^{\pi_{\beta}})$ to ensure compatibility between synthetic and offline data, and a behavior alignment regularizer $D_{f}(\rho_{T}^{\pi} \Vert \rho_{M}^{\pi})$ that maintains consistent policy performance across dynamics models $M$ and real environments $T$. 
        
        While objective~\eqref{eq-surrogate objective} theoretically addresses the synthetic-to-real distribution mismatch (as shown later in Sec.~\ref{bsec-conservative value estimation}), it is impractical to compute due to the inaccessibility of distribution $\rho_{T}^{\pi}$. 
        To address this problem, we apply the convex conjugate theorem to the dual formulation of the RL objective, reformulating it as a maximin optimization problem. 
        This reformulation forms the core of DAMO, a unified maximin framework for mitigating the OOD issue. 
        \begin{theorem}
            \label{thm-final_objective}
            \eqref{eq-surrogate objective} is equivalent to the following problem:
            \begin{equation}
                \max\limits_{\pi} \min\limits_{Q(s,a)} (1-\gamma)              \mathbb{E}_{s\sim\mu_{0},a\sim \pi} [Q(s,a)] 
                + \alpha \mathbb{E}_{\rho_{M}^{\pi}} [ f_{\star}(\Phi(s,a,s')/\alpha) ],
                \label{eq-final_objective}
            \end{equation}
            
        where $f_{\star}(x)$ is the conjugate function of the convex function $f(x)$, and $\Phi(s, a, s') = r - \alpha \log\frac{\rho_{M}^{\pi}}{\rho_{T}^{\pi_{\beta}}} + \gamma \sum\limits_{a'} Q(s', a') \pi(a'|s') - Q(s, a)$
        \end{theorem}
        The proof of Theorem~\ref{thm-final_objective} utilizes the DICE to correct the discrepancies between distributions and can be found in Appendix~\ref{appendix-proof of thm1}.

        \subsection{Practical Implementation of DAMO}
        \label{bsec-Practical algorithm for DAMO}

        To implement the maximin optimization~\eqref{eq-final_objective} practically, we employ a classifier to approximate the data alignment term $\log(\rho_{M}^{\pi} /  \rho_{T}^{\pi_{\beta}})$. 
        We train a classifier $h(s, a, s^{\prime})$ with the following loss function to distinguish transitions sampled from offline data and synthetic data:
        \begin{equation}
            \min_{h} \dfrac{1}{|\mathcal{D}_{R}|} \sum_{(s,a,s') \in \mathcal{D}_{R}} \log h(s,a,s') 
            \\
            + \dfrac{1}{|\mathcal{D}_{M}|} \sum_{(s,a,s') \in \mathcal{D}_{M}} [ \log (1 - h(s,a,s')) ].
            \label{eq-classifier_loss}
        \end{equation}
        $\mathcal{D}_{R}$ and $\mathcal{D}_{M}$ represent the buffer storing offline and synthetic data, respectively. Based on the learned classifier $h^{\star}(s, a, s^{\prime})$, $\log(\rho_{M}^{\pi}/\rho_{T}^{\pi_{\beta}})$ can be computed using the following analytical expression:
        \begin{equation}
            \log\dfrac{\rho_{M}^{\pi}}{\rho_{T}^{\pi_{\beta}}} = \log \dfrac{h^{\star}(s,a,s')}{1 - h^{\star}(s,a,s')}.
            \label{eq-data alignment term}
        \end{equation}

        Note that the maximin optimization problem~\eqref{eq-final_objective} can be decomposed into two phases: the inner value estimation and the outer policy improvement, resembling the actor-critic framework. 
        Thus, we adopt the actor-critic structure to facilitate the implementation of DAMO. The complete framework is listed in Appendix~\ref{appendix-Algo}.


        \subsection{Theoretical Insights for DAMO}
        \label{bsec-conservative value estimation}

        The core structure of DAMO involves mapping the inner minimization in \eqref{eq-final_objective} to the estimation of a dual conservative value $Q(s, a)$, and the outer maximization to policy improvement based on the estimated conservative value. 
        In this subsection, we present deeper theoretical insights into the working mechanism of DAMO.

        \quad
        
        \noindent
        \textbf{Dual Conservative Value Estimation.}
        We first show that the inner minimization within \eqref{eq-final_objective} includes both: (1) an explicit penalty that directly aligns the synthetic data with offline data, and (2) an implicit penalty that emerges during minimization, ensuring policy behavior consistency as the solution converges to the optimum.   
        \begin{theorem}
            \label{thm-proposition}
            The optimal solution $Q_{\pi}^{\star}$ for the inner minimization optimization in \eqref{eq-final_objective} satisfies:
            \begin{equation*}
                Q_{\pi}^{\star} = r - \alpha \log \dfrac{\rho^{\pi}_{M}}{\rho^{\pi_{\beta}}_{T}} - \alpha f^{\prime} ( \dfrac{\rho^{\pi}_{T}}{\rho^{\pi}_{M}}) + \mathcal{T}^{\pi} Q_{\pi}^{\star}.
            \end{equation*}
        \end{theorem}
        The proof of Theorem~\ref{thm-proposition} is provided in Appendix~\ref {appendix-proof of thm2}. This theorem indicates that DAMO inherently introduces an explicit data alignment regularizer, $\log (\rho^{\pi}_{M} / \rho^{\pi{\beta}}_{T})$, and an implicit behavior alignment adjustment, $f^{\prime} (\rho^{\pi}_{T} / \rho^{\pi}_{M})$. 
        The inner minimization thus reshapes the vanilla reward $r(s, a, s^{\prime})$ into a refined reward $\Tilde{r}(s,a,s')$:
        \begin{equation*}
            \Tilde{r} = r - \alpha \log\frac{\rho_{M}^{\pi}}{\rho_{T}^{\pi_{\beta}}} - \alpha f^{\prime} ( \dfrac{\rho^{\pi}_{T}}{\rho^{\pi}_{M}}).
        \end{equation*}
        The data alignment regularizer $\log (\rho^{\pi}_{M} / \rho^{\pi{\beta}}_{T})$ penalizes transition $(s, a, s^{\prime})$ that exhibit discrepancies between synthetic and offline data, promoting alignment with in-distribution transitions.
        This alignment helps ensure accurate value estimation for these transitions.
        Meanwhile, the behavior alignment adjustment $f^{\prime} (\rho^{\pi}_{T} / \rho^{\pi}_{M})$ focuses on that different dynamics impact the distribution of $(s, a, s^{\prime})$. 
        By penalizing these discrepancies, we encourage the agent to select policies that exhibit similar performance in both the dynamics model and the underlying environment, thus mitigating the impact of OOD states during deployment and ensuring consistent policy behavior between dynamics models and real environments.
       
    \quad

    \noindent
    \textbf{Consistent Policy Improvement.}
        Unlike conventional offline actor-critic frameworks that optimize policies solely by maximizing estimated values, DAMO ensures the objective consistency between value estimation and policy improvement. 
        Specifically, DAMO employs the conservative Q-value to minimize the objective~\eqref{eq-final_objective} and utilizes $\pi$ to maximize it. 
        Prior actor-critic approaches typically follow a bilevel optimization framework, while DAMO employs a maximin paradigm, ensuring objective consistency.
        As demonstrated in Sec.~\ref{bsec-Alignment-Consistency Experiments}, this objective-consistency maintains the conservatism of value estimation during policy improvement.
        
        Once the inner minimization is solved, the outer maximization reduces to the surrogate objective~\eqref{eq-surrogate objective}, providing a lower bound for the standard RL objective~$\mathcal{J}(\pi) = \mathbb{E}_{\pi} [\sum_{t=0}^{\infty} \gamma^{t} r(s_{t}, a_{t}, s_{t+1})]$:
        \begin{theorem}
            \label{thm-lower bound}
        	The surrogate objective~\eqref{eq-surrogate objective} serves as a lower bound of $\mathcal{J}(\pi)$ for all $\pi$:
        	\begin{equation*}
        		\mathcal{J}(\pi) \geq \mathbb{E}_{\rho_{T}^{\pi}} [ r(s, a, s^{\prime})-\alpha\log\dfrac{\rho_{M}^{\pi}}{\rho_{T}^{\pi_{\beta}}} ]
        		-\alpha D_{f}(\rho_{T}^{\pi} \Vert \rho_{M}^{\pi}).
        	\end{equation*}
        \end{theorem}
        The detailed proof of Theorem~\ref{thm-lower bound} can be found in Appendix~\ref{appendix-proof of thm3}. This theorem demonstrates the equivalence between maximin objective~\eqref{eq-final_objective} and the standard RL objective, thereby offering a theoretical guarantee that the policy learned through objective~\eqref{eq-final_objective} can achieve consistent performance in the real environment.

\section{Experiments}
\label{sec-experiments}
    In this section, we focus on the following aspects: 1) The capability of DAMO to handle distribution mismatch, and the contributions of each component of DAMO. 2) The performance of DAMO against existing methods on standard benchmarks. 3) The impact of hyperparameter settings and implementation approaches. 4) The OOD generalization ability of DAMO in the task with relabeled data.  

    We delve into these aspects using the D4RL~\cite{fu2020d4rl} and NeoRL~\cite{qin2022neorl} benchmark on the MuJoCo simulator~\cite{todorov2012mujoco}. Our implementation is based on the OfflineRL-Kit library\footnote{https://github.com/yihaosun1124/OfflineRL-Kit}. The basic parameters of DAMO are consistent with the settings of the library. The experiment details and the hyperparameter configuration is listed in Appendix~\ref{appendix-Experiment Setting}.

    \begin{figure}[h]
        \centering
        \subfigure[Normalized Score]{
        \begin{minipage}[b]{0.22\textwidth}
            \centering
            \includegraphics[width=\textwidth]{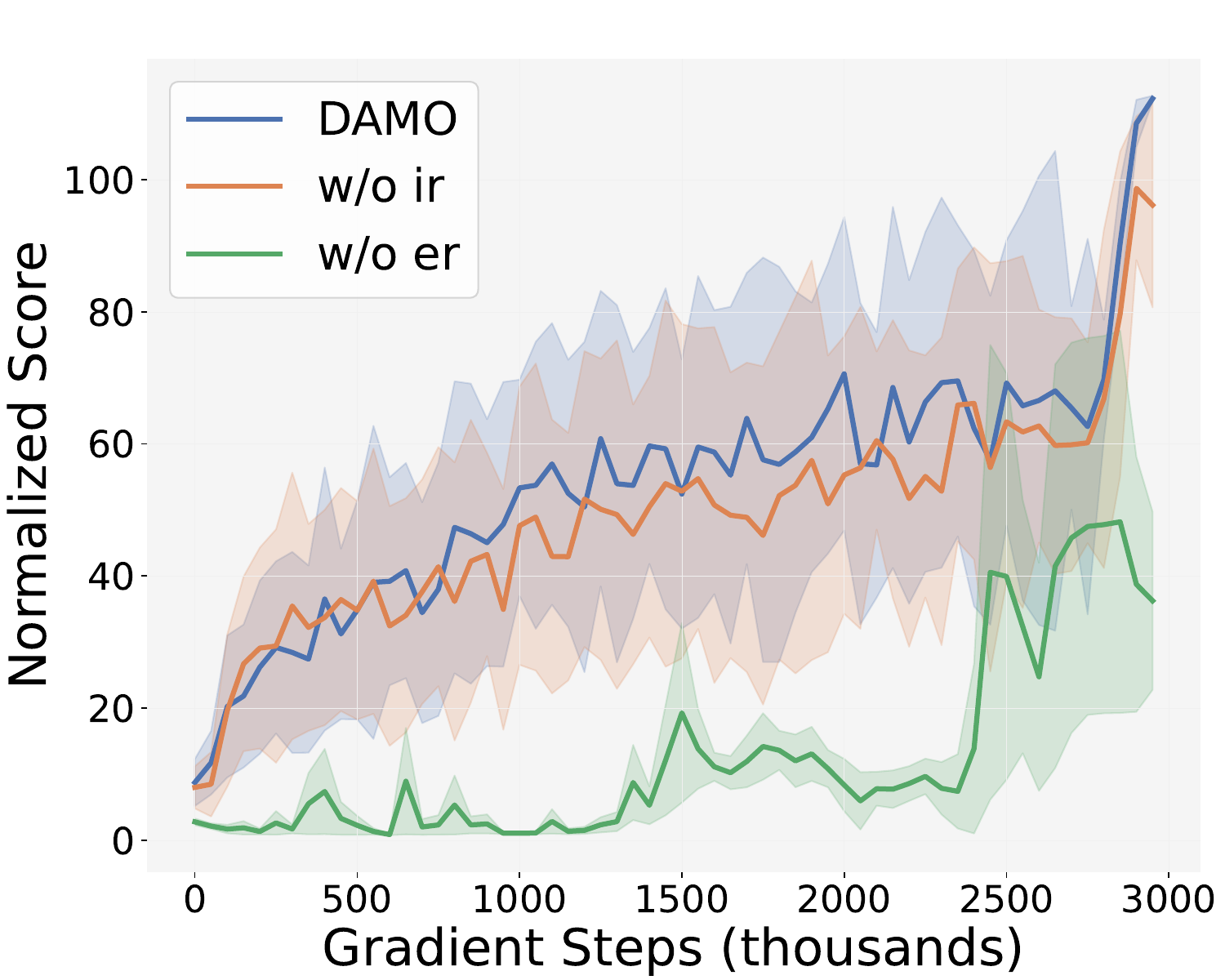}
            \label{subfig-score}
        \end{minipage}
        }
        \subfigure[Estimated Value]{
        \begin{minipage}[b]{0.22\textwidth}
            \centering
            \includegraphics[width=\textwidth]{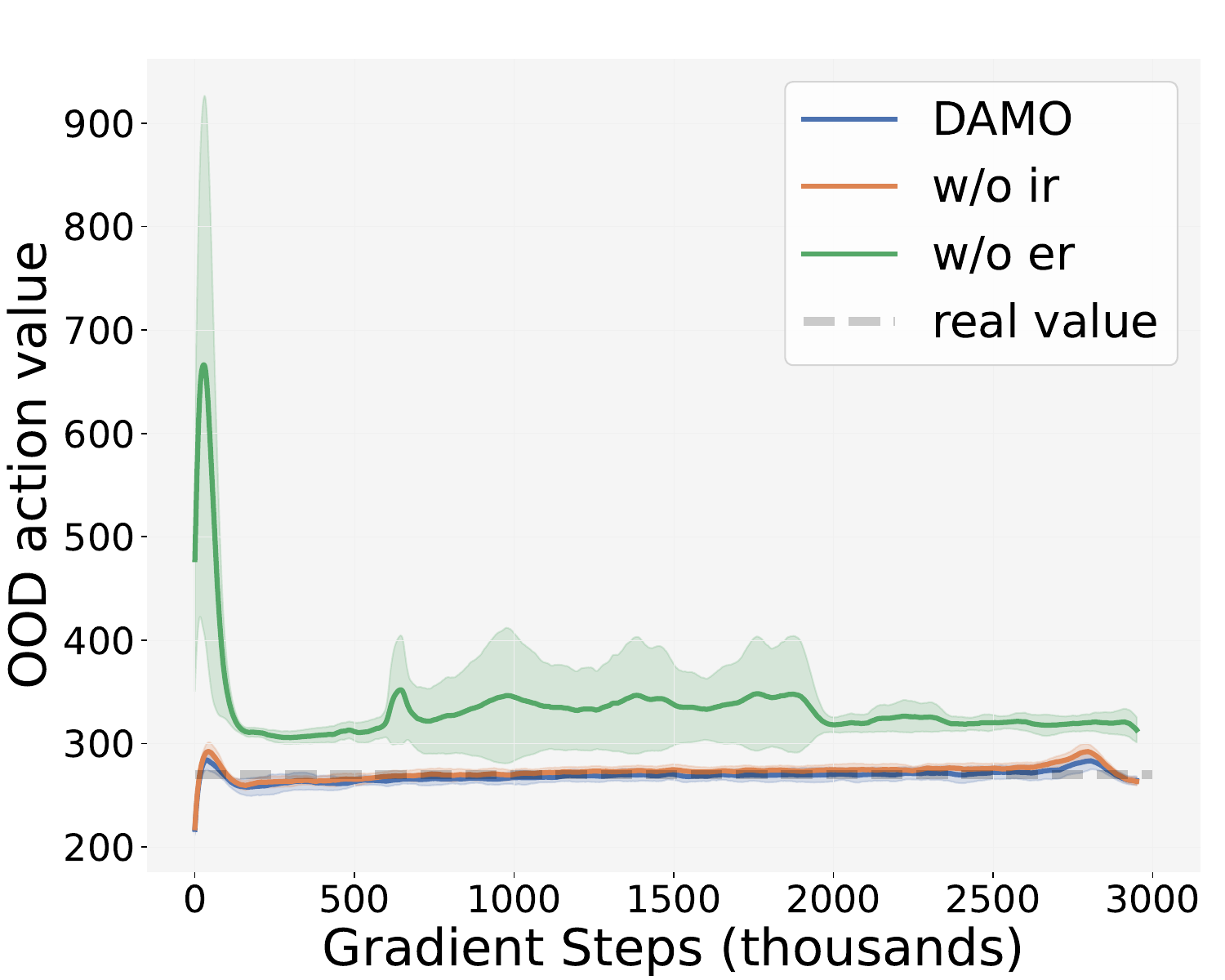}
            \label{subfig-value}
        \end{minipage}
        }
        \caption{Training process of DAMO, w/o ir, w/o er. (a) The data alignment term is the primary factor that influences policy improvement, while the absence of the behavior alignment term results in performance degradation in the final stages. (b) The lack of the data alignment term leads to the overestimation of the value.}
        \label{fig-exp-training process}
    \end{figure}
    \subsection{Effectiveness of DAMO Paradigm}
    \label{bsec-Alignment-Consistency Experiments}
        In this section, we present empirical evidence demonstrating the dual capabilities of DAMO: 1) effectively aligning synthetic data with offline data, and 2) consistently maintaining policy stability between the dynamics models and real environments. Our detailed ablation study thoroughly investigates two key components of DAMO: inner value estimation and outer policy improvement.
        
        \subsubsection{Dual Conservative Value Estimation}

        \quad
        
\begin{figure}[h]
            \centering
            \subfigure{
            \begin{minipage}[b]{0.4\textwidth}
                \centering
                \includegraphics[width=\textwidth]{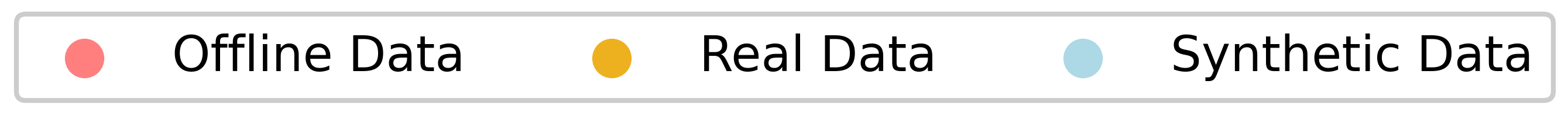}
            \end{minipage}
            }
            \\
            \setcounter{subfigure}{0}
            \subfigure[The distribution of w/o er]{
            \begin{minipage}[b]{0.4\textwidth}  
                \centering
                \includegraphics[width=\textwidth]{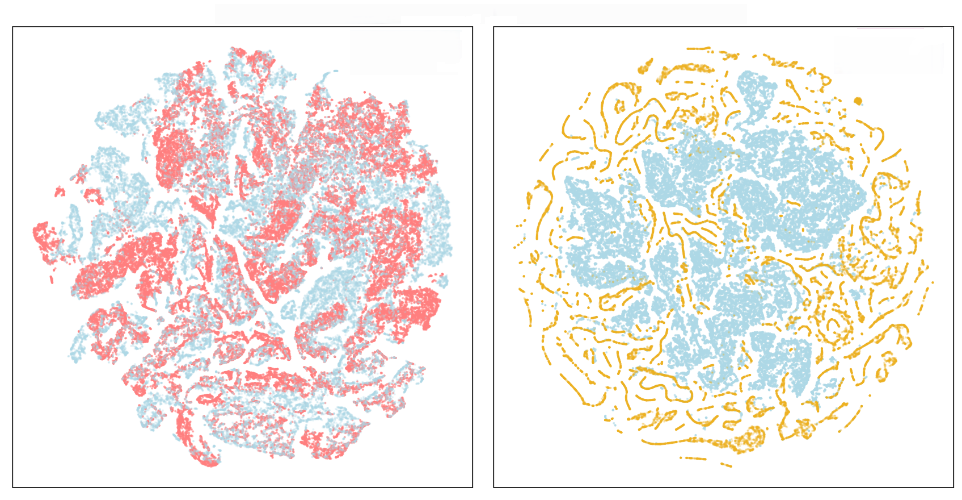}
                \label{subfig-wo_er}
            \end{minipage}
            }
            \subfigure[The distribution of w/o ir]{
            \begin{minipage}[b]{0.4\textwidth} 
                \centering
                \includegraphics[width=\textwidth]{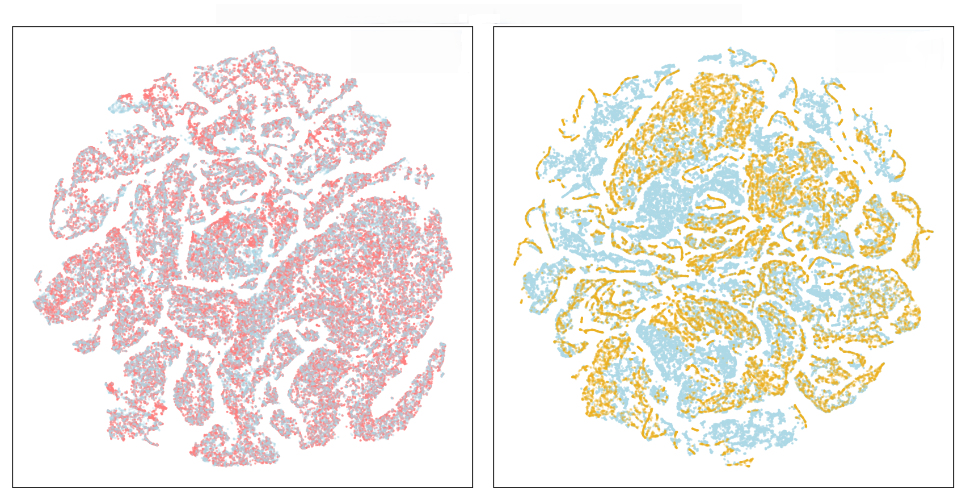}
                \label{subfig-wo_ir}
            \end{minipage}
            }
            \subfigure[The distribution of DAMO]{
            \begin{minipage}[b]{0.4\textwidth}  
                \centering
                \includegraphics[width=\textwidth]{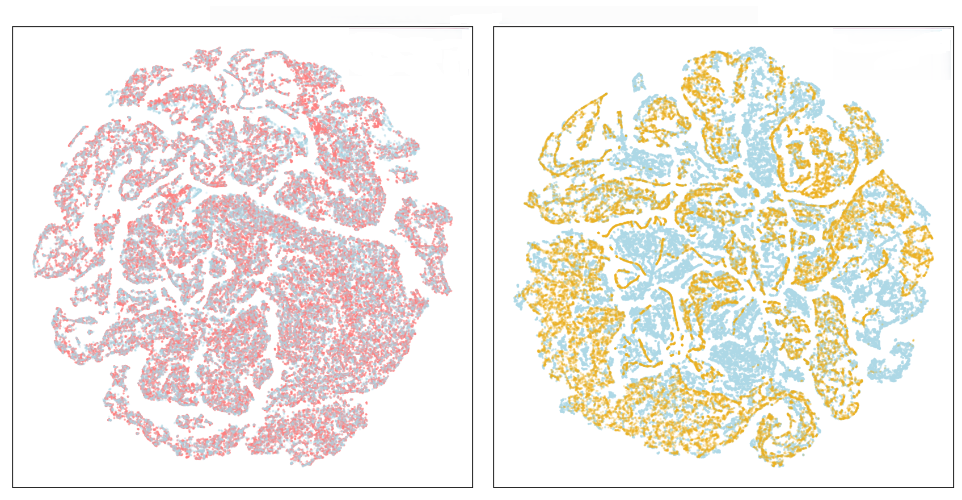}
                \label{subfig-DAMO}
            \end{minipage}
            }
            \caption{(a)-(c) illustrate the results of DAMO without the data alignment term~(w/o er), DAMO without the behavior alignment term~(w/o ir), and standard DAMO. Left panel: state-action (s, a) distributions (offline vs. synthetic). Right panel: state distributions $s$ (real vs. synthetic).} 
            \label{fig-exp-visualization}
        \end{figure}

        \noindent
        In this part, we present an ablation study examining the contributions of the data alignment regularizer and behavior alignment regularizer. We evaluate their impact by removing each term from DAMO and training policies on the hopper-medium-expert task. Specifically, we test DAMO under three distinct conditions: the configuration without the data alignment regularizer (w/o er), the configuration without the behavior alignment regularizer (w/o ir), and the complete DAMO framework. The empirical results are presented in Fig.~\ref{fig-exp-training process}. To further provide intuitive insights, we present the visualization of distributions under each setting in Fig~\ref{fig-exp-visualization}.

        \begin{table*}[t]
            \centering
            \caption{Results on the D4RL Gym benchmark. The numbers reported for DAMO are the normalized scores averaged over the final iteration of training across 4 seeds, with $\pm$ standard deviation. Since we haven't obtained the results of O-DICE on random datasets, we didn't include it in the table and used a dash (-) to indicate the absence of scores. The top three scores are highlighted in bold. We use an asterisk (*) for the best score, an underscore (\_) for the second, and bold text for the third.}
            \label{tab-normalized-score}
            \resizebox{\textwidth}{!}{
            \begin{tabular}{l|ccc|ccccc|c}
                \toprule  Task Name  & Algae &  Opti  & O-DICE & MOPO & COMBO & TT  & RAMBO & MOBILE & DAMO (Ours) 
                \\ \midrule
                halfcheetah-random   & -0.3  & 11.6   & -      & \textbf{38.5} & \underline{\textbf{38.8}}  &  6.1  & $\textbf{39.5}^{*}$  & 38.3  & $34.0 \pm 3.0$
                \\ 
                hopper-random        & 0.9   & 11.2   & -      & $\textbf{31.7}^{*}$ & 17.9  & 6.9  & 25.4  & \textbf{25.5}  & \underline{\textbf{31.5 $\pm$ 0.4}}
                \\ 
                walker2d-random      & 0.5   & \underline{\textbf{9.9}}    & -      & \textbf{7.4}  & 7.0   &  5.9  & 0.0   & $\textbf{21.6}^{*}$  & 0.0
                \\ \midrule
                halfcheetah-medium   & -2.2  & 38.2   & 47.4   & \underline{\textbf{72.4}} & 54.2  &  46.9 & $\textbf{77.9}^{*}$  & 69.3  & \textbf{71.3 $\pm$ 0.5}
                \\ 
                hopper-medium        & 1.2   & \textbf{94.1}   & 86.1   & 62.8 & \underline{\textbf{97.2}}  &  67.1 & 87.0  & $\textbf{102.7}^{*}$ & $93.8 \pm 8.2$
                \\ 
                walker2d-medium      & 0.3   & 21.8   & \underline{\textbf{84.9}}   & 84.1 & 81.9  &  81.3 & \underline{\textbf{84.9}}  & $\textbf{88.7}^{*}$  & $83.1 \pm 8.0 $
                \\ \midrule
                halfcheetah-medium-replay & -2.1 &39.8& 44.0   & $\textbf{72.1}^{*}$ & 55.1  &  44.1 & \textbf{68.7}  & \underline{\textbf{71.4}}  & 61.6 $\pm$ 3.0
                \\ 
                hopper-medium-replay & 1.1   & 36.4   & $\textbf{99.9}^{*}$   & 92.7 & 89.5  &  99.4 & \underline{\textbf{99.5}}  & 57.7  & $\textbf{99.9} \pm \textbf{0.9}^{*}$
                \\ 
                walker2d-medium-replay & 0.6& 21.6   & 83.6   & \textbf{85.9} & 56.0  &  82.6 & \underline{\textbf{89.2}}  & 80.5  & \textbf{89.4} $\pm$ $\textbf{3.0}^{*}$
                \\ \midrule
                halfcheetah-medium-expert &-0.8& 91.1 & 93.2   & 83.6 & 90.0  &  95.0 & \textbf{95.4}  & \underline{\textbf{101.4}} & \textbf{103.1} $\pm$ $\textbf{0.9}^{*} $
                \\ 
                hopper-medium-expert   & 1.1   & \underline{\textbf{111.5}} & 110.8  & 74.6 & \textbf{111.0} & 110.0 & 88.2  & 107.7 & \textbf{112.3} $\pm$ $\textbf{3.6}^{*}$
                \\ 
                walker2d-medium-expert & 0.4   & 74.8 & \underline{\textbf{110.8}}  & 108.2& 103.3 &  101.9& 56.7  & $\textbf{113.0}^{*}$ & \textbf{109.4 $\pm$ 0.8}
                \\ \midrule
                average score  & 0.0 & 43.4 & - & \textbf{68.7} & 66.0 & 61.1 & 66.1 & \underline{\textbf{72.2}} & $\textbf{74.4}^{*}$
                \\ \bottomrule
            \end{tabular}
            }
        \end{table*}

        \quad

        \noindent
        \textbf{Effectiveness of Data Alignment. }
        A comparative analysis between Fig.~\ref{subfig-wo_er} and Fig.~\ref{subfig-DAMO} reveals that w/o er fails to establish alignment between synthetic and offline data, thereby inadequately mitigating OOD state-action pairs. The results in Fig.~\ref{subfig-wo_ir} further highlight that the data alignment regularizer enables the w/o ir configuration to align synthetic data with offline data. To clarify the detrimental effects of incompatibility between these two data sources, the result in Fig.~\ref{subfig-value} illustrates that the w/o er configuration overestimates the value of OOD actions, leading to suboptimal training performance. This is corroborated by the poor evaluation score of the w/o er configuration depicted in Fig.~\ref{subfig-score}.
        
        \quad

        \noindent
        \textbf{Effectiveness of Behavior Alignment. }\quad
        A comparison between Fig.~\ref{subfig-wo_ir} and Fig.~\ref{subfig-DAMO} reveals that while w/o ir successfully achieves synthetic-offline data compatibility and mitigates OOD actions, the absence of the behavior alignment term leads to incomplete mitigation of OOD states, manifesting policy inconsistency between dynamics models and real environments. Although the behavior alignment term imposes constraints, the results in Fig.~\ref{subfig-wo_er} demonstrate that w/o er fails to effectively mitigate OOD states. The observed discrepancies in state distributions between real environment data and model-generated data under the w/o er highlight the critical importance of establishing robust compatibility between synthetic and offline data. This compatibility is essential for ensuring consistent policy performance across dynamics models and real environments. The influence is reflected in Fig.\ref{subfig-score}, which shows performance degradation of w/o ir in the final stage of training. 
        
        \subsubsection{Towards Consistent Policy Improvement} 

        \quad
        
        \noindent
        This section presents a statistical analysis of the value estimation performance of DAMO across multiple datasets, with comparisons against the true value function of the policy in the real environment. Our results validate that the compatible policy improvement framework effectively maintains the conservatism of value estimation.

        To investigate the significance of objective alignment in DAMO, we conduct a comparison experiment by directly integrating DAMO with standard off-policy mechanisms~(Inconsistent Version) while evaluating the critic's estimation of state-action pair values and comparing it with the original DAMO~(Consistent Version). The detailed implementation of the Inconsistent Version can be found in Appendix~\ref{appendix-inconsistent version of DAMO}. We perform this analysis on the medium-expert dataset across three distinct environments: HalfCheetah, Hopper, and Walker2d. The obtained quantitative results are summarized in Table~\ref{tab-policy improvement}.
        The experimental findings reveal that the Consistent Version achieves conservative value estimations across all environments. Although the Inconsistent version maintains conservatism in the HalfCheetah and Hopper environments, it exhibits significant overestimation in the Walker2d environment. These results demonstrate the critical role of objective alignment in persisting conservative value estimation and preventing overestimation.
        
        \begin{table}
            \centering
            \caption{The average value of OOD state-action pairs, $\infty$ for severe overestimation. The smallest values are bolded, which indicates the best mitigation of the overestimation problem. }
            \label{tab-policy improvement}
            \resizebox{0.5\textwidth}{!}{
            \begin{tabular}{l|ccc}
                \toprule 
                                & halfcheetah & hopper & walker2d
                \\ \midrule
                Consistent Version  &   \textbf{95.2}      & \textbf{213.0}  & \textbf{231.7}
                 \\ 
                 Inconsistent Version &   101.4     & 219.4  & $\infty$
                 \\
                 $Real \ value$ &   819.6     & 269.6  & 299.7
                 \\ \bottomrule
            \end{tabular}
            }
            \vspace{-2em}
        \end{table}
    \subsection{Comparison Results}
        \begin{table*}[t]
            \centering
            \caption{Results on the NeoRL benchmark. We report the normalized scores averaged over the trained policy across 4 seeds. The results of MOBILE are obtained by executing the official implementation.}
            \label{tab-normalized-score-neorl}
            \resizebox{\textwidth}{!}{
            \begin{tabular}{l|*{4}{c}|*{2}{c}|c}
                \toprule  Task Name & BC & CQL & TD3+BC & EDAC & MOPO & MOBILE & DAMO (Ours) 
                \\ \midrule
                Halfcheetah-L  & 29.1 & 38.2 & 30.0 & 31.3 & 40.1 & \textbf{54.8} & $47.55 \pm 3.11$
                \\ 
                Hopper-L       & 15.1 & 16.0 & 15.8 & 18.3 & 6.2  & 18.2  & \textbf{24.05 $\pm$ 2.18}
                \\ 
                Walker2d-L     & 28.5 & 44.7 & 43.0 & 40.2 & 11.6 & 23.7 & $\textbf{46.59} \pm \textbf{9.18}$
                \\ \midrule
                Halfcheetah-M  & 49.0 & 54.6 & 52.3 & 54.9 & 62.3 & \textbf{79.0} & $69.04 \pm 1.13$
                \\ 
                Hopper-M       & 51.3 & 64.5 & \textbf{70.3} & 44.9 & 1.0 & 43.4 & $64.57 \pm 9.17$
                \\ 
                Walker2d-M     & 48.7 & 57.3 & 58.6 & 57.6 & 39.9 & 60.1 & $\textbf{66.49} \pm \textbf{2.28} $
                \\ \midrule
                Halfcheetah-H  & 71.3 & 77.4 & 75.3 & \textbf{81.4} & 65.9 & 71.8 & $67.23 \pm 17.5$
                \\ 
                Hopper-H       & 43.1 & \textbf{76.6} & 75.3 & 52.5 & 11.5 & 42.3 & $74.87 \pm 6.39$
                \\ 
                Walker2d-H     & 72.6 & 75.3 & 69.6 & \textbf{75.5} & 18.0 & 71.9   & $74.28 \pm 2.03$
                \\ \bottomrule
            \end{tabular}
            }
        \end{table*}
    
        \subsubsection{D4RL}

        \quad

        \noindent
        \textbf{Dataset.} We first evaluate the performance of DAMO using the D4RL~\cite{fu2020d4rl}. Our assessments cover 12 distinct datasets across three control environments (HalfCheetah, Hopper, and Walker2d) and four dataset categories: random, medium, medium-replay, and medium-expert. Following standard evaluation protocols, we utilize the "v2" versions of all datasets in D4RL for consistent comparison.

        \quad
        
        \noindent
        \textbf{Baseline.} We compare DAMO with two types of offline RL baselines: DICE-based algorithms, which follow a similar form of training objective to DAMO, and Model-based algorithms, which achieve SOTA performance in D4RL benchmark. For DICE-based approaches, Algae-DICE~\cite{algaedice} transforms the intractable state-action-level constraint into a unified objective for policy training. Opti-DICE~\cite{optidice} directly estimates the stationary distribution corrections of the optimal policy to attain a high-rewarding policy. O-DICE~\cite{odice} uses the orthogonal gradient update to diminish the conflict of different gradients during policy training. For model-based approaches, MOPO~\cite{mopo} penalizes rewards via the predicted variance of ensembled dynamics models. COMBO~\cite{combo} implements CQL within a model-based framework. TT~\cite{TT} uses a transformer to model offline trajectories and employs beam search for planning. RAMBO~\cite{rambo} adversarially trains the policy and the dynamics model within a robust framework. MOBILE~\cite{mobile} penalizes the value learning of synthetic data based on the estimated uncertainty of the Bellman Q-function, which is derived from an ensemble of models.

        \quad
        
        \noindent
        \textbf{Comparison Results.} Table~\ref{tab-normalized-score} reports the scores in the D4RL benchmark. Overall, DAMO demonstrates superior performance, achieving the highest average score across all baseline methods. Notably, for high-quality datasets such as medium-expert and medium-replay, DAMO consistently attains SOTA or near-SOTA performance. These results suggest that effective dual alignment, combined with a high-quality behavior policy, enable the agent to learn high-reward policies within ID regions. Furthermore, DAMO maintains strong performance on medium-quality datasets. This demonstrates the robustness of DAMO in handling less-than-ideal data conditions and highlights the importance of ensuring both model-environment policy consistency and synthetic-offline data compatibility. However, the performance of DAMO on random datasets reveals limitations to some extent, indicating the need for more aggressive exploration strategies in low-quality data scenarios to identify uncovered regions. This phenomenon is pronounced in the Walker2d-random dataset, where DAMO struggles to learn an effective policy. One possible explanation is that the optimization landscape of the Walker2d contains numerous suboptimal solutions corresponding to saddle points in the maximin problem. If this is the case, such structural characteristics could potentially trap DAMO in local optima, limiting its ability to discover better solutions. Further investigation is needed to confirm the above hypothesis.

        \subsubsection{NeoRL}

        \begin{table*}[t]
            \centering
            \caption{Average returns on halfcheetah-jump. Batch Mean and Batch Max denote the mean and maximum average return across all trajectories in the offline dataset, respectively.}
            \label{tab-average return of halfcheetah-jump}
            \resizebox{\linewidth}{!}{
            \begin{tabular}{l|*{2}{c}|*{3}{c}|*{3}{c}}
                 \toprule
                 Environment & Batch Mean & Batch Max & DAMO & MOPO & COMBO & BEAR & BRAC-v & TD3+BC
                 \\ \midrule
                 halfcheetah-jump & -1184 & 2887 & $4871 \pm 416$ & $5030 \pm 435$ & $4868\pm 175$ & $186\pm 231$ & $1038 \pm 179$ & $1340 \pm 303$
                 \\ \bottomrule
            \end{tabular}
            }
        \end{table*}

        \quad

        \noindent
        \textbf{Dataset.} To further assess the performance of DAMO in real-world scenarios, we test it in the NeoRL~\cite{qin2022neorl}. By emulating real-world data-collection scenarios, the datasets in NeoRL characterize a narrow scope and limited coverage, which impose significant challenges for offline RL. Our assessments include nine datasets across three environments (HalfCheetah, Hopper, Walker2d) and three dataset quality levels (L, M, H), corresponding to low, medium, and high.

        \quad
        
        \noindent
        \textbf{Baseline.} We have not chosen the DICE-based method for comparison, since the results of these methods can not be found in the original paper or the NeoRL paper, and searching for optimal hyperparameters for each method would take excessive time. Instead, we select SOTA offline model-free methods to ensure the diversity of the baseline. For model-based approaches, we keep two representative methods, MOPO~\cite{mopo} and MOBILE~\cite{mobile} for comparison. For model-free approaches, BC~\cite{bain1995framework} imitates the behavior policy from the offline dataset. CQL~\cite{kumar2020conservative} penalizes the Q-values of OOD samples by incorporating regularization into the value loss function. TD3+BC~\cite{fujimoto2021minimalist} introduces a behavior cloning regularization into the TD3 objective. EDAC~\cite{an2021uncertainty} penalizes the Q-function based on uncertainty derived from an ensemble of Q-networks.

        \quad

        \noindent
        \textbf{Comparison Results.} Table~\ref{tab-normalized-score-neorl} reports the scores in the NeoRL benchmark. DAMO maintains competitive performance under more realistic and difficult scenarios, indicating strong potential for real-world applicability. Specifically, the strong generalization ability in narrow data coverage induced by suboptimal policy facilitates DAMO in achieving SOTA or near-SOTA performance on datasets of low-quality and medium-quality. Among high-quality datasets, DAMO secures a near-SOTA score in the Walker2d-H dataset and a high ranking score in the Hopper-H dataset. However, while training with the Halfcheetah-H dataset, DAMO fails to learn the optimal in-distribution policy, which contradicts our analysis in D4RL results. Through inspecting the training process, we observe that although DAMO achieves rapid policy improvement during the early phase, the subsequent training process exhibits significant fluctuations, with occasional steep drops in performance. We attribute this instability to limited data coverage caused by the near-deterministic nature of the policy and the noise in the environment, which together present a substantial challenge for robust learning.

    \subsection{OOD generalization of DAMO}

    Model-based methods are generally more capable of performing OOD generalization than model-free methods. As DAMO follows the idea of behavior regularization, which potentially hinders the agent from exploration, it remains unclear whether DAMO preserves the superior generalization ability~\cite{lyu2022mildly, liu2024adaptive}. To clarify this problem, we evaluate DAMO in the halfcheetah-jump task~\cite{mopo} where the agent solves a task that entails the effective generalization outside the scope of the offline data. Specifically, a behavior policy is trained online to control the cheetah to move forward as fast as possible. The rewards of the collected trajectories are then reshaped to encourage the cheetah jumping during offline training. Since the behavior policy is trained for a different purpose, the dataset only contains the suboptimal policy for jumping. Therefore, OOD generalization is crucial for the agent to acquire a high-return policy. We compare DAMO with other model-based methods and behavior-regularization methods. The implementation of the experiment and introduction of baselines are presented in Appendix~\ref{appendix-The concrete setting of halcheetah-jump}. As shown in Table~\ref{tab-average return of halfcheetah-jump}, DAMO achieves comparable performance with model-based methods and outperforms all behavior-regularization methods. This suggests that combining MBRL with behavior regularization enables the agent to achieve both offline conservatism and OOD generalization when handling OOD issues in offline RL.
        
    \begin{figure}[h]
        \centering
        \subfigure[Tuning of $\alpha$]{
        \begin{minipage}{0.22\textwidth}
            \centering
            \includegraphics[width=\textwidth]{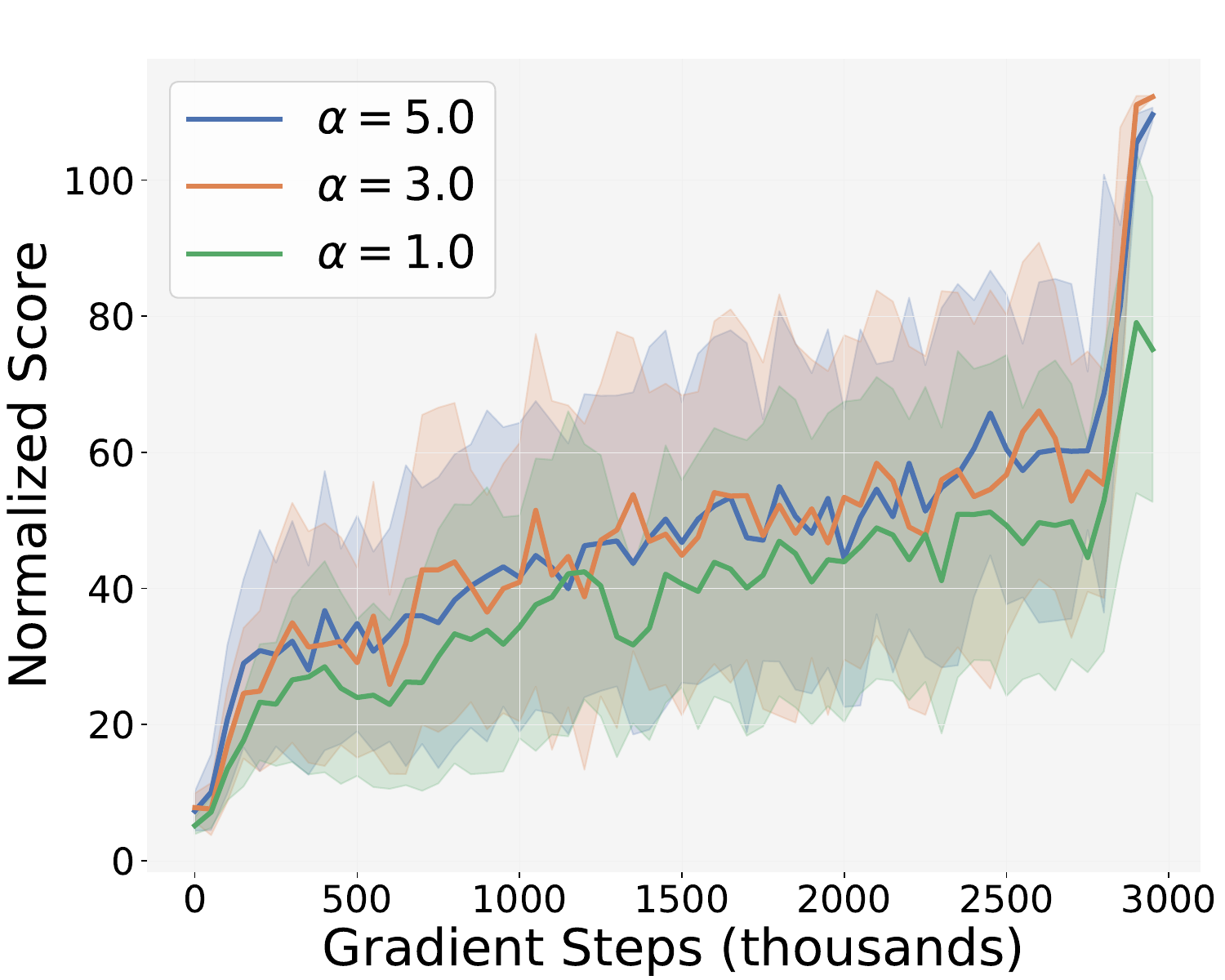}
            \label{subfig-hyperparameter}
        \end{minipage}
        }
        \subfigure[DAMO without fixing $\alpha$]{
        \begin{minipage}{0.22\textwidth}
            \centering
            \includegraphics[width=\textwidth]{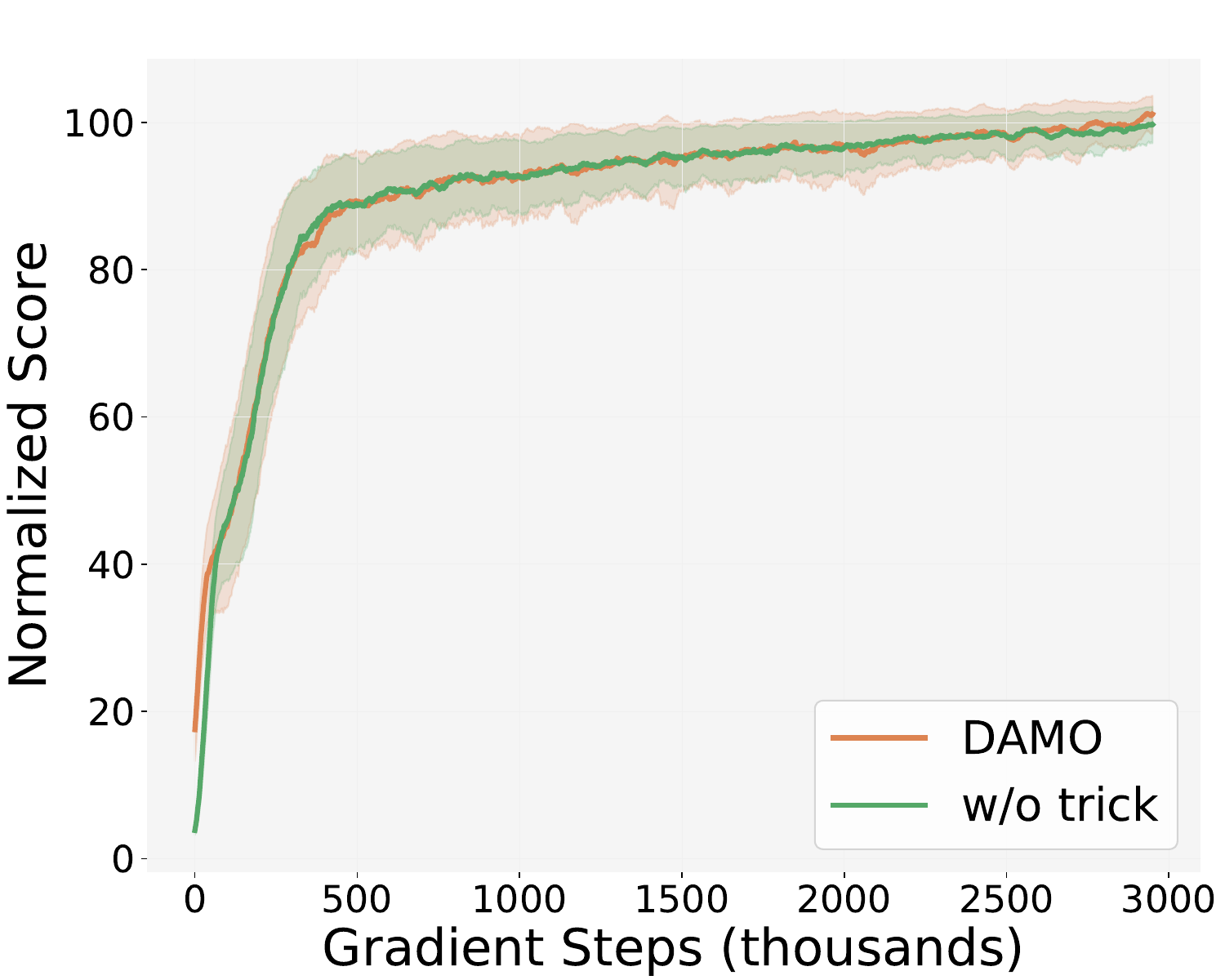}
            \label{subfig-trick}
        \end{minipage}
        }
        \caption{(a) The tuning experiment of $\alpha$ was performed on the hopper-medium-expert dataset for validation. (b) The fixing experiment of $\alpha$ was performed on the halfcheetah-medium-expert dataset for validation.}
    \end{figure}
    \subsection{Effects of Coefficient $\alpha$}
    \label{bsec-hyperparameter tunning}
        DAMO is regulated by a key hyperparameter: the hyperparameter $\alpha$, which controls the extent of policy discrepancies.
        While we consider $\alpha$ as a hyperparameter controlling the degree of conservatism, the empirical results in Fig.~\ref{subfig-hyperparameter} demonstrate that DAMO maintains robust performance across a wide range of $\alpha$ values, with its primary impact observed in the final stage performance. DAMO achieves superior performance with larger $\alpha$ values (e.g., $\alpha=3.0,5.0$), whereas insufficiently small values (e.g., $\alpha=1.0$) tend to promote excessive risk-taking during exploration, failing to converge to an optimal policy.
        To enhance training stability, we initially follow the prior work~\cite{luo2024ompo} and fix $\alpha$ to 1.0 in the actor training. We present the concrete implementation in Appendix~\ref{appendix-modification of actor training objective}. However, experimental results demonstrate that DAMO achieves competitive performance even without this specific implementation. To further validate this observation, a comparative experiment is conducted, evaluating both the standard DAMO implementation and its variant without fixed $\alpha$ under their optimal hyperparameter configurations. As shown in Fig.~\ref{subfig-trick}, both settings achieve competitive performances, confirming that DAMO maintains robust performance regardless of the specific implementation.

\section{Conclusion and Discussion}
\label{sec-Conclusion and Discussion}
    This paper examines limitations in existing offline MBRL approaches, highlighting the crucial need to address behavior inconsistency caused by policy discrepancies between biased models and real environments. 
    Building upon our analysis, we introduce a unified framework that concurrently ensures both synthetic-offline data compatibility and model-environment policy consistency, with experimental validation demonstrating its effectiveness. 
    Significantly, the maximin optimization framework of DAMO demonstrates the benefits of aligning the actor and critic training objectives within the actor-critic architecture, ensuring essential conservatism during policy updates. Our future work could extend DAMO to areas such as offline-to-online RL~\cite{li2023proto}, where the method with adaptability and the capability to bridge the synthetic-to-real distribution gap is in great demand. The limitations of DAMO are focused on two-fold: the sampling of the initial state and the computation of the data alignment term. Although we assume the distribution of the initial state is accessible, some offline datasets do not explicitly label the initial states. Therefore, we randomly sample a state from these offline datasets to serve as the initial state, which creates a deployment gap. A classifier is employed to assist computation. However, the accuracy of the classifier is critical: imprecise classifiers fail to estimate the modified reward, leading to unstable policy training and potentially invalidating the objective as a lower bound.



\bibliographystyle{ACM-Reference-Format} 
\bibliography{sample}
\nocite{*}


\appendix
\section{Proofs in the Main Text}
\label{appendix-proof}
    We present a detailed proof for each theorem discussed above. The sequence of proofs follows the logic of derivation rather than strictly adhering to their order of appearance in the main text.

    \subsection{Proof of Theorem~\ref{thm-lower bound}}
    \label{appendix-proof of thm3}

    To prove Theorem~\ref{thm-lower bound}, we first introduce an upper bound of the KL-divergence.
    \begin{lemma}
    \label{lemma-1}
        For any convex function $f(x)$ satisfying the inequality: $f(x) \geq x\log x, \forall x>0$, the KL-divergence admits the following upper bound:
        \begin{equation*}
                D_{KL}(\rho_{T}^{\pi} \Vert \rho_{T}^{\pi_{\beta}})  \leq \mathbb{E}_{\rho_{T}^{\pi}} [ \log\dfrac{\rho_{M}^{\pi}}{\rho_{T}^{\pi_{\beta}}} ] 
            + D_{f}(\rho_{T}^{\pi} \Vert \rho_{M}^{\pi}) .
        \end{equation*}
        \label{thm-KL upper bound}
        \begin{proof}
    	\begin{align*}
    		D_{KL}(\rho_{T}^{\pi} \Vert \rho_{T}^{\pi_{\beta}}) & = \mathbb{E}_{\rho_{T}^{\pi}} [ \log\dfrac{\rho_{T}^{\pi}}{\rho_{T}^{\pi_{\beta}}} ]
    		\\
    		& = \mathbb{E}_{\rho_{T}^{\pi}} [ \log\dfrac{\rho_{T}^{\pi}}{\rho_{T}^{\pi_{\beta}}} 
    		\cdot \dfrac{\rho_{M}^{\pi}}{\rho_{M}^{\pi}} ]
    		\\
    		& = \mathbb{E}_{\rho_{T}^{\pi}} [ \log\dfrac{\rho_{M}^{\pi}}{\rho_{T}^{\pi_{\beta}}} 
    		] + D_{KL}(\rho_{T}^{\pi} \Vert \rho_{M}^{\pi})
    		\\
    		& \leq \mathbb{E}_{\rho_{T}^{\pi}} [ \log\dfrac{\rho_{M}^{\pi}}{\rho_{T}^{\pi_{\beta}}} 
    			] + D_{f}(\rho_{T}^{\pi} \Vert \rho_{M}^{\pi})
    	\end{align*}
        To establish the final inequality, we invoke the property that the f-divergence dominates the KL divergence, i.e., $D_{f}(\rho_{T}^{\pi} \Vert \rho_{M}^{\pi}) \geq D_{KL}(\rho_{T}^{\pi} \Vert \rho_{M}^{\pi})$, which stems from the inequality $f(x)\geq x\log x$ imposed on the convex function $f(x)$.
        \end{proof}
    \end{lemma}
    
    Deploying the Lemma~\ref{thm-KL upper bound} into the objective $\max\limits_{\pi} \mathbb{E}_{\rho_{T}^{\pi}} [ r(s, a, s^{\prime}) ] - \alpha D_{KL}(\rho_{T}^{\pi} \Vert \rho_{T}^{\pi_{\beta}})$, then we have a lower bound of vanilla RL objective $\mathcal{J}(\pi) = \mathbb{E}_{\rho_{T}^{\pi}} [ r(s,a,s^{\prime}) ]$.
    \begin{corollary}
        We have the following surrogate objective that serves as a lower bound of $\mathcal{J}(\pi)$:
	\begin{equation*}
		  \mathcal{J}(\pi) \geq \mathbb{E}_{\rho_{T}^{\pi}} [ r(s,a,s^{\prime})  -  \alpha\log\dfrac{\rho_{M}^{\pi}}{\rho_{T}^{\pi_{\beta}}} 
			] - \alpha D_{f}(\rho_{T}^{\pi} \Vert \rho_{M}^{\pi}). \quad \forall \alpha \geq 0
	\end{equation*}
        \begin{proof}
            The following objective is a lower bound of $\mathcal{J}(\pi)$ due to the non-negativity of KL-divergence:
            \begin{equation*}
                \mathbb{E}_{\rho_{T}^{\pi}} [ r(s, a, s^{\prime}) ] - \alpha D_{KL}(\rho_{T}^{\pi} \Vert \rho_{T}^{\pi_{\beta}})
            \end{equation*}
            We substitute the $D_{KL}(\rho_{T}^{\pi} \Vert \rho_{T}^{\pi_{\beta}})$ with its upper bound in theorem~\ref{thm-KL upper bound}, then for any policy $\pi$, we have:
            \begin{equation*}
                \mathcal{J}(\pi) \geq \mathbb{E}_{\rho_{T}^{\pi}} [ r(s,a,s^{\prime})  -  \alpha\log\dfrac{\rho_{M}^{\pi}}{\rho_{T}^{\pi_{\beta}}} ] - \alpha D_{f}(\rho_{T}^{\pi} \Vert \rho_{M}^{\pi}).
            \end{equation*}
        \end{proof}
    \end{corollary}
    We have now completed the proof of Theorem~\ref {thm-lower bound}. 
    
    \subsection{Proof of Theorem~\ref{thm-final_objective}}
    \label{appendix-proof of thm1}
    Next, we present the proof of Theorem~\ref{thm-final_objective}. We first introduce two lemmas.

    Firstly, we define the conjugate functional and provide a corresponding proposition of biconjugate.
    \begin{lemma}
        \label{thm-biconjugate function}
        Given a functional $f: \mathbb{E} \rightarrow \mathbb{R}$, $\mathbb{E}$ is a Hilbert space. Its  Fenchel conjugate functional $f_{\star}: \mathbb{E} \rightarrow \mathbb{R}$ is defined as:
        \begin{equation*}
            f_{\star}(y) = \sup\limits_{x\in \mathbb{E}}\{ \langle y,x \rangle - f(x) \}
        \end{equation*}
         $\langle y,x \rangle$ is the inner dot of $y,x$. We call the functional $f_{\star \star} = (f_{\star})_{\star}$ as the biconjugate functional of $f$. If $f$ is a convex, closed functional, then its biconjugate functional is itself:
        \begin{equation*}
            f_{\star \star} = f
        \end{equation*}
        \begin{proof}
            The proof of the proposition about the biconjugate functional can be found in \cite{fenchel2014conjugate}
        \end{proof}
    \end{lemma}
    Then, in the second lemma, we consider the convex functional.
    \begin{lemma}
        \label{thm-variational form of f-divergence}
		We can represent f-divergence in a functional form with $x(z)$: $D_{f}(x \Vert p) = \mathbb{E}_{z\sim p} f(\frac{x(z)}{p(z)}) \triangleq h_{p}(x)$, that means we fix the distributional function $p$ and treat $D_{f}(x \Vert p)$ as a functional $h_{p}(x)$. Then its Fenchel conjugate function is $\mathbb{E}_{z\sim p} f_{\star}(y(z))$
        \begin{proof}
    	We use $g_{\star}(y)$ to denote the conjugate functional of $D_{f}(x \Vert p)$, $\mathcal{X}$ to denote the $L^{2} space$. Then we have:
    	\begin{align*}
    		g_{\star}(y) &= \sup\limits_{x\in \mathcal{X}} \{\langle y,x \rangle - h_{p}(x)  \}
    		\\
    		&= \sup\limits_{x\in \mathcal{X}}\{ \int_{R}y(z)x(z) dz - \int_{R} p(z)\cdot f(\frac{x(z)}{p(z)})dz \}
    		\\
    		&= \sup\limits_{x\in \mathcal{X}}\{ \int_{R} p(z)[y(z)\frac{x(z)}{p(z)} -  f(\frac{x(z)}{p(z)}) ]dz \}
    		\\
    		&= \int_{R} p(z)\sup\limits_{x(z)}[y(z)\frac{x(z)}{p(z)} -  f(\frac{x(z)}{p(z)}) ]dz
    		\\
    		&= \int_{R} p(z)\sup\limits_{x'(z)}[y(z)x'(z) -  f(x'(z)) ]dz
    		\\
    		&= \int_{R} p(z)f_{\star}(y(z))dz
    		\\
    		&= \mathbb{E}_{z\sim p} f_{\star}(y(z))
    	\end{align*}
        The first equality is the definition of the Fenchel conjugate function. In the second equality, we give the explicit expressions for the inner product and the KL divergence. In the fifth equality, we make the variable change and substitute $(x(z)/ p(z))$ with $x'(z)$. The sixth equality follows the definition of the Fenchel conjugate function.
        \end{proof}
    \end{lemma}
    We can employ Lemma~\ref{thm-biconjugate function}, \ref{thm-variational form of f-divergence} to represent the f-divergence in surrogate objective~\eqref{eq-surrogate objective} in a MAX-form.
    \begin{corollary}
    \label{thm-f-divergence in max form}
    The f-divergence $D_{f}(\rho_{T}^{\pi} \Vert \rho_{M}^{\pi})$ satisfy
	\begin{equation*}
		D_{f}(\rho_{T}^{\pi} \Vert \rho_{M}^{\pi}) = \max\limits_{y(s,a,s')} \mathbb{E}_{\rho_{T}^{\pi}} [ y(s,a,s') ] -
	\mathbb{E}_{\rho_{M}^{\pi}} [ f_{\star}(y(s,a,s')) ] .
	\end{equation*}
        \begin{proof}
            We can verify that $D_{f}(x \Vert p)$ is a convex, closed functional. Then the biconjugate functional of $D_{f}(x \Vert p)$ is itself according to Lemma~\ref{thm-biconjugate function}. We keep the notation and use $g_{\star}(y)$ to denote the conjugate function of $D_{f}(x \Vert p)$, $y(s, a, s')$ is any function defined in the transition space. Then $D_{f}(x \Vert p)$ is the conjugate of $g_{\star}(y)$. Therefore, $ D_{f}(\rho_{T}^{\pi} \Vert \rho_{M}^{\pi})$ can be represented by the conjugate functional of $g_{\star}(y)$:
            \begin{align*}
                D_{f}(\rho_{T}^{\pi} \Vert \rho_{M}^{\pi}) &= \max\limits_{y} \langle y, \rho^{\pi}_{T} \rangle - g_{\star} (y)
                \\
                &= \max\limits_{y} \int \rho^{\pi}_{T} \cdot y \ \mathrm{d}s \ \mathrm{d}a \ \mathrm{d}s'  - g_{\star} (y)
                \\
                &= \max\limits_{y} \mathbb{E}_{\rho_{T}^{\pi}} [ y ] -\mathbb{E}_{\rho_{M}^{\pi}} [ f_{\star}(y) ] 
            \end{align*}
            The first equality is the definition of the Fenchel conjugate function. In the second equality, we express the inner product in the form of an integral. In the third equality, we represent $g_{\star}(y)$ in the form of expectation according to Lemma~\ref{thm-variational form of f-divergence}.
        \end{proof}
    \end{corollary}
    With the Corollary~\ref{thm-f-divergence in max form}, we can attain a maximin form of surrogate objective~\eqref{eq-surrogate objective}.
    \begin{theorem}
        Optimizing surrogate objective~\eqref{eq-surrogate objective} is equivalent to solving the following maximin problem
        \begin{equation}
        \label{eq-first maximin form}
		\max\limits_{\pi} \min\limits_{y} \mathbb{E}_{\rho_{T}^{\pi}} [ r(s,a,s^{\prime})  -  \alpha\log\dfrac{\rho_{M}^{\pi}}{\rho_{T}^{\pi_{\beta}}} 
			- \alpha y] + \alpha\mathbb{E}_{\rho_{M}^{\pi}} [f_{\star}(y)]
	  \end{equation}
        \label{thm-the first maximin form}
        \begin{proof}
            Substituting the $D_{f}(\rho_{T}^{\pi} \Vert \rho_{M}^{\pi})$ in surrogate objective~\eqref{eq-surrogate objective} with its MAX-form in Corollary~\ref{thm-f-divergence in max form}, then we attain the equivalent maximin form of surrogate objective~\eqref{eq-surrogate objective}.
        \end{proof}
    \end{theorem}
    
    Finally, we should make the maximin objective feasible in the offline RL setting, which means that we should eliminate the expectation over $\rho^{\pi}_{T}$ in \eqref{eq-first maximin form}.
    \begin{theorem}
    The maximin objective~\eqref{eq-first maximin form} is equivalent to the following objective
	\begin{equation}
           \max\limits_{\pi}\min\limits_{Q(s,a)} (1-\gamma) \mathbb{E}_{s \sim \mu_{0},a \sim \pi}           [Q(s,a)] + \alpha \mathbb{E}_{\rho_{M}^{\pi}} f_{\star}(\Phi(s,a,s')/\alpha)
            \label{eq-maximin}
	\end{equation}
        \label{thm-maximin}
        \begin{proof}
        Defining $y(s,a,s') = \frac{1}{\alpha}r(s,a,s^{\prime}) - \alpha \log\frac{\rho_{M}^{\pi}}{\rho_{T}^{\pi_{\beta}}} + \gamma \mathbb{E}_{a^{\prime}\sim\pi}Q(s^{\prime},a^{\prime}) - Q(s,a)$. For the convenience, we define $\beta_{t}(s,a,s')$
    	\begin{equation*}
    		\beta_{t}(s,a,s') = Pr[s_{t} = s, a_{t} = a, s_{t+1} = s'|s_{0}\sim \mu_{0} , a_{t}\sim \pi , s_{t+1}\sim T]
    	\end{equation*}
        According to definition, $\rho_{T}^{\pi}(s,a,s') = (1-\gamma)\sum\limits_{t=0}^{\infty} \gamma^{t}\beta_{t}$.
        Similarly, we can define $\beta_{t}(s, a)$ and $\beta_{t}(s,a,s',a')$.
        \begin{align*}
            & \beta_{t}(s,a) = Pr[s_{t} = s, a_{t} = a|s_{0}\sim \mu_{0} , a_{t}\sim \pi , s_{t}\sim T]
            \\
            & \beta_{t}(s,a,s',a') = Pr[s_{t} = s, a_{t} = a, s_{t+1} = s', a_{t+1} = a'|\mu_{0} ,\pi ,T]
        \end{align*}
        Substituting $y(s,a,s')$ with $[ r(s,a,s^{\prime}) - \alpha \log\frac{\rho_{M}^{\pi}}{\rho_{T}^{\pi_{\beta}}} + \gamma \mathbb{E}_{a^{\prime}\sim\pi}Q(s^{\prime},a^{\prime}) - Q(s,a) ] / \alpha$, then the first term in objective~\eqref{eq-first maximin form} can be represented as:
    	\begin{align*}
                & \mathbb{E}_{\rho_{T}^{\pi}} [ r(s,a,s^{\prime})  -  \alpha\log\dfrac{\rho_{M}^{\pi}}{\rho_{T}^{\pi_{\beta}}} 
			- \alpha y(s,a,s')]
                    \\
    		= & \mathbb{E}_{\rho_{T}^{\pi}} [ Q(s,a) - \gamma \mathbb{E}_{a^{\prime}\sim\pi}Q(s^{\prime},a^{\prime}) ] 
    			\\
    		= & \sum\limits_{s,a,s'} \rho_{T}^{\pi}(s,a,s') [Q(s,a) - \gamma\sum\limits_{a'}Q(s',a') \pi(a'|s')] 
    			\\
    			= &  \sum\limits_{s,a,s'} [\rho_{T}^{\pi}(s,a,s')Q(s,a)] - \sum\limits_{s,a,s',a'} [\gamma \rho_{T}^{\pi}(s,a,s') Q(s',a') \pi(a'|s')] 
    			\\
    		= & (1-\gamma) [ \sum\limits_{s,a,s'} \sum\limits_{t=0}^{\infty} \gamma^{t} \beta_{t}(s,a,s')Q(s,a)  - \sum\limits_{s,a,s',a'} \sum\limits_{t=0}^{\infty} \gamma^{t+1}\beta_{t}(s,a,s')Q(s',a')\pi(a'|s') ]
    			\\
    		= & (1-\gamma) [ \sum\limits_{s,a,s'} \sum\limits_{t=0}^{\infty} \gamma^{t}
    			\beta_{t}(s,a,s')Q - \sum\limits_{s,a,s',a'} \sum\limits_{t=0}^{\infty} \gamma^{t+1}\beta_{t}(s,a,s',a')Q(s',a')  ]
    			\\
    		= & (1-\gamma) [ \sum\limits_{s,a}\sum\limits_{t=0}^{\infty} \gamma^{t}
    			\beta_{t}(s,a)Q - \sum\limits_{s',a'} \sum\limits_{t=0}^{\infty} \gamma^{t+1}\beta_{t}(s',a')Q(s',a') ] 
    			\\
    		= & (1-\gamma) \sum\limits_{s,a}\beta_{0}(s,a)Q(s,a)
    			\\
    		= & (1-\gamma) \mathbb{E}_{s \sim \mu_{0},a\sim \pi}[Q(s,a)] 
    	\end{align*}
        In the first equality, we make a variable change of $y(s, a, s')$. In the second equality, we explicitly formulate the operator $\tau^{\pi}$. In the third equality, we employ the proposition $\sum\limits_{s, a, s'} \rho_{T}^{\pi}(s,a,s') = 1$. In the fourth equality, we express $\rho_{T}^{\pi}(s,a,s')$ in the discounted sum of $\beta_{t}(s,a,s')$. In the fifth equality, we employ the proposition $\beta_{t}(s,a,s') = \beta_{t}(s,a,s', a') \pi(a'|s') $. In the sixth equality, we employ the proposition $\sum\limits_{s, a} \beta_{t}(s,a,s',a') = \beta(s',a')$.
        
        For the second term in objective~\eqref{eq-first maximin form}, we define $\Phi(s, a, s') = \alpha y(s, a, s')$, then objective~\eqref{eq-first maximin form} can be written as:
        \begin{equation*}
           \max\limits_{\pi}\min\limits_{Q(s,a)} (1-\gamma) \mathbb{E}_{s \sim \mu_{0},a \sim \pi} [Q(s,a)] + \alpha \mathbb{E}_{\rho_{M}^{\pi}} f_{\star}(\Phi(s,a,s')/\alpha)
	\end{equation*}
        \end{proof}
    \end{theorem}

    We have now completed the proof of Theorem~\ref {thm-final_objective}. 
    
    \subsection{Proof of Theorem~\ref{thm-proposition}}
    \label{appendix-proof of thm2}
    Next, we present the proof of Theorem~\ref{thm-proposition}.
    \begin{theorem}
        If $Q^{\star}(s,a)$ is the optimal solution of inner minimization in maximin problem~\eqref{eq-maximin}, then it satisfies 
        \begin{equation}
            Q^{\star}(s,a) = r(s,a,s^{\prime}) - \alpha \log \dfrac{\rho^{\pi}_{M}}{\rho^{\pi_{\beta}}_{T}} - \alpha f^{\prime} ( \dfrac{\rho^{\pi}_{T}}{\rho^{\pi}_{M}}) + \mathcal{T}^{\pi} Q^{\star}(s,a)
        \end{equation}
        \label{thm-inner optimal}
        \begin{proof}
            $f(x)$ is a convex function, so its conjugate function $f_{\star}$ is also convex. The objective function below is convex about $y(s,a,s^{\prime})$
            \begin{equation*}
                \max\limits_{\pi} \min\limits_{y(s,a,s')} \mathbb{E}_{\rho_{T}^{\pi}} [ r(s,a,s^{\prime})  -  \alpha\log\dfrac{\rho_{M}^{\pi}}{\rho_{T}^{\pi_{\beta}}} 
			- \alpha y(s,a,s')] + \alpha\mathbb{E}_{\rho_{M}^{\pi}} [f_{\star}(y(s,a,s'))]
            \end{equation*}
            Therefore, the optimality condition is satisfied when the derivative of $y(s, a, s')$ equals zero, and a closed-form solution of $y^{\star}(s, a, s')$ can be obtained: $y^{\star}(s,a,s^{\prime}) = (f^{\prime}_{\star})^{-1} ( \dfrac{\rho^{\pi}_{T}}{\rho^{\pi}_{M}})$. 
            Let $f(x)$ be a strictly convex function,
            then we have $(f^{\prime}_{\star})^{-1} (x) = f^{\prime}(x)$, thus $y^{\star}(s,a,s^{\prime}) = f^{\prime}( \dfrac{\rho^{\pi}_{T}}{\rho^{\pi}_{M}})$.

            Recalling that we employ the variable substitution: $y(s,a,s') = [ r(s,a,s^{\prime}) - \alpha \log\frac{\rho_{M}^{\pi}}{\rho_{T}^{\pi_{\beta}}} + \mathcal{T}^{\pi} Q(s,a) - Q(s,a) ] / \alpha$, so we have
            \begin{equation*}
                y^{\star}(s,a,s') = \dfrac{ r(s,a,s^{\prime}) - \alpha \log\frac{\rho_{M}^{\pi}}{\rho_{T}^{\pi_{\beta}}} + \mathcal{T}^{\pi} Q^{\star}(s,a) - Q^{\star}(s,a)}{\alpha}
            \end{equation*}
            Then the optimal solution $Q^{\star}(s,a)$ satisfies
            \begin{equation*}
                Q^{\star}(s,a) = r(s,a,s^{\prime}) - \alpha \log \dfrac{\rho^{\pi}_{M}}{\rho^{\pi_{\beta}}_{T}} - \alpha f^{\prime} ( \dfrac{\rho^{\pi}_{T}}{\rho^{\pi}_{M}}) + \mathcal{T}^{\pi} Q^{\star}(s,a)
            \end{equation*}
        \end{proof}
    \end{theorem}

    \section{OOD Issues under Dynamics Model Integration}
    \label{app-bsec-OOD actions and states in offline model-based RL}

        OOD issues in offline model-based RL are intrinsically linked to the mismatch between synthetic and real data distributions, compounded in model-based settings, where model inaccuracies introduce additional sources of error.

        \subsection{OOD Actions}
        The dynamics model generates data that can significantly deviate from the distribution of the offline dataset, leading to OOD actions.
        This triggers negative model exploitation~\cite{levine2020offline}, resulting in poor policy performance. 
        Aligning the distributions of synthetic and offline data can effectively mitigate these issues. 
        The model, $M$, is used to generate synthetic data for training, but it is essential to avoid taking OOD actions when rolling out policy in the model, to ensure accurate value estimation.
        
        Specifically, in an offline setting, where interaction with the real environment is not possible, overestimating the value of OOD actions leads to model exploitation.
        Consequently, the agent may learn policies that perform poorly in the underlying environment compared to the model, leading to algorithmic failure during deployment.
        To avoid model exploitation, it is crucial to align synthetic with offline data.  
        Given offline data collected under the behavior policy $\pi_{\beta}$,  We aim to find a policy $\pi$ that behaves in model $M$ like $\pi_{\beta}$ does in environment $T$. 
        Specifically, we aim to match the distribution of $(s, a)_{M}^{\pi}$ with $(s, a)_{T}^{\pi_{\beta}}$, ensuring accurate value estimation of $(s, a)_{M}^{\pi}$. 

        \subsection{OOD states}
        The performance of the learning policy during deployment is critically dependent on the alignment of transition pairs between the learned model and the real environment. 
        Significant discrepancies between these transitions can cause performance degradation and lead to OOD states, highlighting the necessity of correcting transition biases during training.
        Even if the learned policy performs well in the model, it is important, during real-environment deployment, to avoid OOD states, for which we cannot accurately estimate the value function in the offline setting.
        
        For example, if the learned model predicts the transition $(s, a, s_{1})$, but the real environment produces a transition $(s, a, s_{2})$, where $s_{1}$ is a high-value state and $s_{2}$ is a low-value state,  the policy will perform poorly in the real environment.
        The situation worsens if $s_{2}$ is an OOD state that has never been encountered in model $M$ before. 
        In this scenario, the agent cannot estimate the value of the state-action pair $(s_{2}, a^{\prime})$, leading to ineffective decision-making and model-environment policy inconsistency. 
        In an offline setting, since we cannot adjust the policy by interacting with the environment, we must approximate and correct the discrepancies between the model $M$ and the real environment $T$ during training.
        This ensures that the distribution of transition $(s, a, s^{\prime})_{M}^{\pi}$ matches the distribution of $(s, a, s^{\prime})_{T}^{\pi}$ in the underlying environment.

        In summary, while optimizing the policy $\pi$ to maximize returns, we should constrain it in two directions: 1) aligning the synthetic data $(s, a, s^{\prime})_{M}^{\pi}$ with the offline data $(s, a, s^{\prime})_{T}^{\pi_{\beta}}$, and 2) ensuring that $\pi$ exhibits consistent behavior in both dynamics models and real environments.

\begin{algorithm}[h]
    \caption{DAMO}
    \label{alg-DAMO}
    \begin{algorithmic}[1]
        \Require Offline dataset $\mathcal{D}_{R}$, dynamics model $M$, actor network $\pi_{\phi}$, critic network $Q_{\theta}$, classifier $h_{\psi}$.
        \State Train the probabilistic dynamics model $M$ on $\mathcal{D}_{R}$
        \State Initialize $\mathcal{D}_{M} \leftarrow \varnothing$
        \For{each epoch}
            \State Generate synthetic rollouts by model $M$
            \State Add transitions to $\mathcal{D}_{M}$
            \State Update $h_{\psi}$ via Eq.~\eqref{eq-classifier_loss}
            \State Calculate $\log(\rho_{M}^{\pi}/\rho_{T}^{\pi_{\beta}})$ via Eq.~\eqref{eq-data alignment term}
            \State Update $Q_{\theta}$ by minimizing Eq.~\eqref{eq-final_objective}
            \State Update $\pi_{\phi}$ by maximizing Eq.~\eqref{eq-final_objective}
        \EndFor
    \end{algorithmic}
\end{algorithm}
\section{Experiment Details}

\label{appendix-Implementation Details}
    In this section, we present the detailed implementation of DAMO. 

    \subsection{Algorithm of DAMO}
    \label{appendix-Algo}
    
    The complete algorithm of DAMO is summarized in Algorithm~\ref{alg-DAMO}. Two replay buffers, $\mathcal{D}_{R}$ and $\mathcal{D}_{M}$, are used to store transitions from the offline dataset and model rollouts, respectively. 
    
    \subsection{Selection of f-divergence}
        As established in Theorem~\ref{thm-lower bound}, the f-divergence should consistently exceed the KL-divergence to guarantee that the objective in Equation~\eqref{eq-final_objective} serves as a lower bound for the standard RL objective~$\mathcal{J}(\pi)$. We adhere to the selection made in the prior work~\cite{luo2024ompo} and specifically choose $f(x) = \frac{1}{3}(x - 1)^{3}$, its conjugate function is $f_{\star}(x) = \frac{2}{3}(x-1)^{\frac{3}{2}}.$ While for most values, $f(x)$ exceeds $x\log x$, in a small interval around $x=1$, $f(x)$ fails to statisfy the condition. In principle, it is feasible to choose a f-divergence satisfying the given condition, but due to the tight schedule, we haven't tried other choices of f-divergence. However, we suppose that the choice of f-divergence won't enormously affect the effectiveness of DAMO, since f-divergence primarily serves as a measure of discrepancy. 

        \begin{table}[h]
            \centering
            \caption{Hyperparameter of DAMO when optimizing policy}
            \label{tab-Hyperparameter of DAMO when optimizing policy}
            \resizebox{\linewidth}{!}{
            \begin{tabular}{ccl}
                \toprule
                 Hyperparameter & Value & Description
                 \\ \midrule
                 $n$      & 7     & Ensemble number of dynamics model
                 \\
                 $n_{elite}$ &  5  & Number of dynamics model for prediction
                 \\
                 $d$      & 0.5   & Sampling ratio for classifier training
                 \\ 
                 $\gamma$ & 0.99  & Discounted factor.
                 \\
                $l_{r}$ of critic & $3\times 10^{-4}$ & Critic learning rate.
                 \\
                 Batch size   &  256  &  Batch size for each update.
                 \\
                 $N_{iter}$   &  3M   &  Total gradient steps.
                 \\ \bottomrule
            \end{tabular}
            }
        \end{table}

    \subsection{The inconsistent version of DAMO}
    \label{appendix-inconsistent version of DAMO}
        Section 6.1.2 focuses on evaluating the effect of DAMO’s consistent policy improvement objective. Prior offline MBRL methods primarily concentrate on the design of conservative Q-functions. Once a conservative Q-function is obtained, these methods typically apply the SAC framework for policy improvement, i.e., 
        \[ \max_{\pi} E_{s \sim D, a\sim \pi}Q^{\pi}(s, a) + \alpha \mathcal{H}(\pi). \]
        where $D$ denotes the offline dataset and $\mathcal{H}(\pi)$ denotes the entropy of the policy $\pi$.
    
        In contrast, DAMO performs policy training directly based on the maximin formulation (2), where the policy improvement objective is consistent with the policy evaluation objective: 
        \[ \max_{\pi}\min_{Q} (1-\gamma) E_{s\sim\mu_{0},a\sim \pi} [Q(s, a)] + \alpha E_{\rho_{M}^{\pi}} [ f_{\star}(\Phi(s, a,s')/\alpha) ]. \]
        We refer to the variant of DAMO that applies the SAC-style policy improvement objective as the “inconsistent version.”

    \subsection{Modification of actor training objective}
    \label{appendix-modification of actor training objective}
        When optimizing the actor network using objective~\eqref{eq-final_objective}, we observed that the substantial residual of the Bellman error often disrupts the optimization of the actor and leads to training instability. To stabilize the training process, we followed the implementation of previous DICE work~\cite{odice, sikchi2023imitation} and fixed the hyperparameter $\alpha$ to $1.0$ during the optimization of the actor network. The specific form of the actor training objective is as follows:
        \begin{equation*}
            \max\limits_{\pi} (1-\gamma)              \mathbb{E}_{s\sim\mu_{0},a\sim \pi} [Q(s,a)] 
                         \\
                + \mathbb{E}_{\rho_{M}^{\pi}} [ f_{\star}(\Phi(s,a,s')) ]
        \end{equation*}
        It should be noted that the effectiveness of DAMO does not rely on this trick. We demonstrated this in the experimental section.\ref{bsec-hyperparameter tunning}.

    \subsection{The concrete setting of halcheetah-jump}
    \label{appendix-The concrete setting of halcheetah-jump}
        \textbf{Data collection.} A SAC agent is trained in the halfcheetah task online for 1M environment steps. The reward function used in this phase is
        \begin{equation*}
            r(s, a) = \max\{v_x, 3 \} - 0.1\times \Vert a \Vert_{2}^{2},
        \end{equation*}
        where $v_x$ denotes the velocity along the x-axis.
        
        \textbf{Reward shaping.} After collecting the offline dataset, we relabel the reward function to:
        \begin{equation*}
            r(s, a) = \max\{v_x, 3\} - 0.1\times \Vert a \Vert_{2}^{2} + 15\times (z- init z),
        \end{equation*}
        where $z$ denotes the z-position of the half-cheetah and init z denotes the initial z-position. As the observation of the original halfcheetah environment does not contain the information of init z, we add an extra dimension to the observation to inform the agent of init z.

        \textbf{Baselines.} Three typical offline behavior-regularization methods are selected for comparison. BEAR~\cite{kumar2019stabilizing}aims to constrain the policy's actions to lie in the support of the behavioral distribution. BRAC~\cite{wu2019behavior} is a family of algorithms that operate by penalizing the value function by some measure of discrepancy (KL divergence or MMD) between $\pi(\cdot|s)$ and $\pi^{B}(\cdot|s)$. TD3+BC~\cite{fujimoto2021minimalist} introduces a behavior cloning regularization into the TD3 objective.

    \subsection{Model, classifier and Policy Optimization}
        For dynamics model $M$, it was depicted as a probabilistic neural network that generates a Gaussian distribution for the subsequent state and reward, contingent upon the current state and action:
        \begin{equation*}
            m_{\theta}(s_{t+1}, r_{t}|s_{t}, a_{t}) = \mathcal{N}(\mu_{\theta}(s_{t}, a_{t}), \Sigma_{\theta}(s_{t}, a_{t})).
        \end{equation*}
        Our model training approach is consistent with the methodology used in prior works~\cite{mopo, mobile}. We train an ensemble of seven dynamics models and select the best five based on their validation prediction error from a held-out set containing 1000 transitions in the offline dataset $D_{R}$. Each model in the ensemble is a 4-layer feedforward neural network with 200 hidden units. We randomly choose one model from the best five models during model rollouts.

        For classifier $h_{\psi}$, we implement the following approach: At each gradient step, we randomly sample data from both offline dataset $D_{R}$ and synthetic dataset $D_{M}$ with a fixed ratio $d$. This method ensures a balanced utilization of data from both $D_{R}$ and $D_{M}$ during training.
    
        Policy optimization is conducted using the Actor-Critic framework in our approach, with hyperparameter configurations aligned with previous offline model-based settings. During each update, a batch of 256 transitions is sampled, where 5\% is derived from the offline dataset $D_{R}$ and the remaining 95\% from the synthetic dataset $D_{M}$. Additionally, We use two common tricks in DAMO: double Q-learning and entropy regularizer. The specific hyperparameter settings applied for the D4RL benchmark are detailed in Table.\ref{tab-Hyperparameter of DAMO when optimizing policy}

\section{Additional Experiments}
\label{appendix-Additional Experiments}

\subsection{A quantitative analysis of Fig~\ref{fig-mopo}}
When drawing Fig~\ref{fig-mopo}, we applied T-SNE to perform dimensionality reduction on the data and visualized the processed data to provide an intuitive understanding of data similarity. During the T-SNE process, we did not quantitatively measure the non-overlapping regions between datasets. To enable a more rigorous comparison, we computed the Wasserstein distance between the datasets. The results are in the Table~\ref{tab-wasserstein-distance}. Experimental results indicate that all three methods achieve a good alignment between model data and offline data, but only DAMO successfully aligns model data with real data, which is consistent with our analysis in the paper. Comparing the directly computed Wasserstein distance with the Wasserstein distance after T-SNE dimensionality reduction, we observe that although the absolute values change, the relative ordering remains the same, confirming the validity of using T-SNE for visualization.

\begin{table}[h]
    \centering
    \caption{Wasserstein distance between the offline, real, and model data. The first column clarifies the type of data, for example, Offline-Model means the Wasserstein distance between Offline data and Model data. 'w/ Tsne' represents the usage of T-SNE to reduce the dimension of data and then calculate the Wasserstein distance. }
    \label{tab-wasserstein-distance}
    \begin{tabular}{l|cc|c}
        \toprule
         Distance& MOPO & MOBILE & DAMO (ours)
         \\ \midrule
         Offline-Model & 0.24 & 0.13 & 0.16
         \\
         Model-Real & 0.39 & 0.43 & 0.27
         \\ \midrule
         Offline-Model(w/ T-SNE) & 2.17 & 1.96 & 2.01
         \\
         Model-Real(w/ T-SNE) & 10.28 & 6.76 & 4.35
         \\ \bottomrule
    \end{tabular}
\end{table}

\subsection{The performance in noisy environments}
\begin{table}[h]
    \centering
    \caption{Results on the noisy mujoco.}
    \label{tab-noisy-env}
    \resizebox{\linewidth}{!}{
    \begin{tabular}{l|cc|c}
        \toprule
         Task Name& MOPO & MOBILE & DAMO (ours)
         \\ \midrule
         Halfcheetah-me & 42.10 & 72.53 & $74.51 \pm 7.55$
         \\
         Hopper-me & 63.82 & 93.86 & $95.36 \pm 4.14$
         \\
         Walker2d-me & 108.69 & 114.90 & $113.94 \pm 1.15$
         \\ \bottomrule
    \end{tabular}
    }
\end{table}
Noisy environment refers to a situation where random noise interferes with the agent’s state observation during testing, which is normal in real-world applications. Therefore, we have supplemented the following experiment in Mujoco to test whether DAMO is robust enough to demonstrate strong performance in noisy environment. We still use DAMO to train the policy on clean offline data, but during the testing phase, at each step transition, the environment introduces noise to the state observed by the agent with a certain probability. Under this setting, we tested the performance of DAMO, MOPO, and MOBILE, and the results are in the table~\ref{tab-noisy-env}. Both DAMO and MOBILE exhibit robust performance in noisy environments, effectively mitigating the impact of environmental noise on policy performance, whereas MOPO experiences a significant performance degradation. Notably, injecting noise in the Walker2d environment does not lead to a noticeable decline in policy performance, which may be attributed to the characteristics of the environment and task, warranting further investigation.

\subsection{DAMO with different hyperparameters for dual alignment}
\begin{figure}
    \centering
    \includegraphics[width=\linewidth]{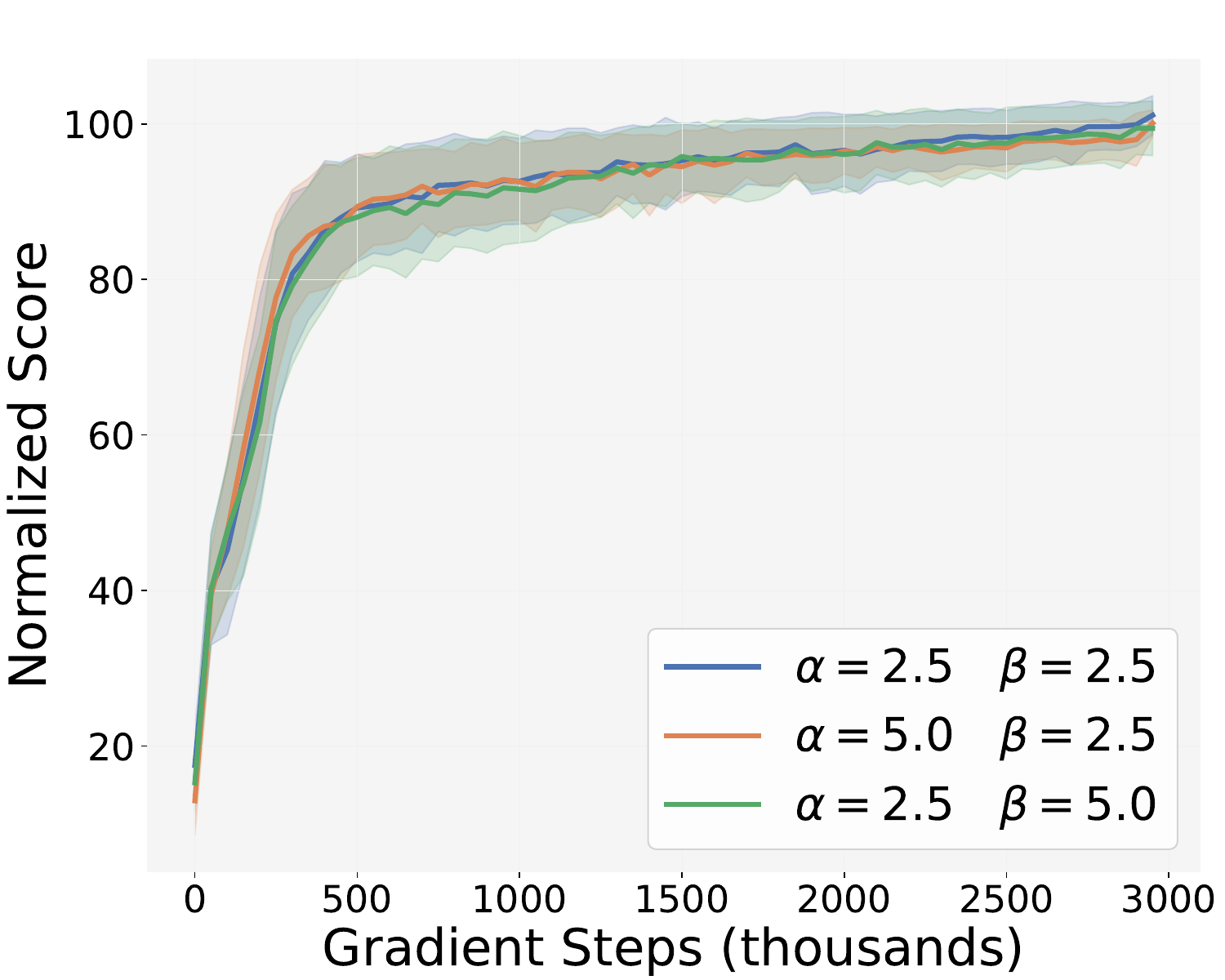}
    \caption{fig-Setting different parameters for two penalization terms}
    \label{fig-different-param}
\end{figure}

Using different parameters for the two regularization terms in the DAMO can serve as a trick to improve algorithm performance. We conducted an experiment under the halfcheetah-medium-expert dataset to evaluate its impact. We set the coefficient of the term $D_{f}(\rho_{T}^{\pi} \Vert \rho_{M}^{\pi})$ in (1) to be $\beta$ and keep the coefficient of the term $\log(\rho_{M}^{\pi} / \rho_{T}^{\pi_{\beta}})$ to be $\alpha$. The training process is illustrated in Fig~\ref{fig-different-param}.
This trick does not yield a significant effect on the Halfcheetah-Medium-Expert dataset. This may be attributed to two factors: DAMO itself is insensitive to the parameters $\alpha$ and $\beta$, and the Halfcheetah environment also exhibits low sensitivity to these parameters. The final scores from top to bottom are 103.1, 101.78, and 100.33.
    
\section{Experiment Setting}
\label{appendix-Experiment Setting}
    \subsection{Benchamark}
    In the D4RL benchmark~\cite{fu2020d4rl}, we focus on the “v2” version of Gym tasks. For DICE-based methods (Algae-DICE, Opti-DICE, and O-DICE), we directly adopt the scores reported in their respective original papers. However, since the results for O-DICE on the random dataset are not provided in the original paper, we exclude them from our comparison. For offline model-based methods, we retrained MOPO~\cite{mopo} and MOBILE~\cite{mobile} on the “v2” Gym datasets. For the remaining methods, we directly report the scores provided in their original papers. All results are summarized in Table~\ref{tab-normalized-score}.

    In the NeoRL benchmark~\cite{qin2022neorl}, we report the performance of BC, CQL, and MOPO based on the original NeoRL paper, while the performance of TD3+BC, EDAC, and MOBILE is derived from the scores reported in the MOBILE~\cite{mobile} paper.

    \begin{table}[h]
        \centering
        \caption{Hyperparameters of DAMO used in the D4RL datasets.}
        \resizebox{\linewidth}{!}{
        \begin{tabular}{lccc}
            \toprule
             Task Name                  & $\alpha$ & $K$ & $l_{r}$ of actor
             \\ \midrule 
             halfcheetah-random         & 0.1 & 5 & $1 \times 10^{-4}$
             \\
             hopper-random              & 2   & 5 & $1 \times 10^{-5}$
             \\
             walker2d-random            & 2   & 5 & $1 \times 10^{-4}$
             \\ \midrule 
             halfcheetah-medium         & 0.5 & 5 & $1 \times 10^{-4}$
             \\
             hopper-medium              & 2.0 & 5 & $1 \times 10^{-5}$
             \\
             walker2d-medium            & 2   & 1 & $5 \times 10^{-5}$
             \\ \midrule 
             halfcheetah-medium-replay  & 0.1 & 5 & $1 \times 10^{-4}$
             \\
             hopper-medium-replay       & 3   & 5 & $1 \times 10^{-4}$
             \\
             walker2d-medium-replay     & 1.5 & 5 & $1 \times 10^{-4}$
             \\ \midrule
             halfcheetah-medium-expert  & 2.5 & 5 & $1 \times 10^{-4}$
             \\
             hopper-medium-expert       & 3   & 5 & $1 \times 10^{-5}$
             \\
             walker2d-medium-expert     & 1.5 & 3 & $1 \times 10^{-4}$
             \\ \bottomrule
        \end{tabular}
        }
        \label{tab-hyperparameter in d4rl}
    \end{table}

    \subsection{Hyperparameters}
    We provide a list of the hyperparameters that were tuned during our experiments. The detailed configurations in the D4RL benchmark are summarized in Table~\ref{tab-hyperparameter in d4rl}. The detailed configurations in the NeoRL benchmark are summarized in Table~\ref{tab-hyperparameter in neorl}. It should be noted that we considered two additional hyperparameters in the NeoRL benchmark, due to the narrow data coverage and noisy environments.

    \textbf{Coefficient $\alpha$.} The coefficient $\alpha$ acts as the sole hyperparameter in our objective function, regulating the divergence between the training policy and the behavior policy. We tune $\alpha$ in the range of $[0.1, 3.0].$

    \textbf{Rollout Length $K$.} Similar to MOPO, we employ short-horizon branch rollouts in DAMO. For Gym tasks in the D4RL benchmark, the horizon parameter $K$ is tuned within the range of $\{ 1, 3, 5\}$.

    \textbf{The actor learning rate $l_{r}$.} Since DAMO operates as a Maximin optimization problem and is susceptible to saddle points, we optimize the actor's learning rate to achieve robust performance. The choice of learning rate plays a pivotal role in DAMO's effectiveness. We tune the actor learning rate $l_r$ in the range of $\{1\times 10^{-5}, 5\times 10^{-5}, 1\times 10^{-4} \}$.

    \textbf{Offline data ratio.} The limited coverage of the transition space leads to poor performance of the trained dynamics in some datasets. Consequently, we have to downweight the ratio of synthethic data used to optimize the objective for stabilizing the training process. We tune the offline data ratio within the range of $\{0.05, 0.1, 0.3\}$.

    \textbf{Auto-alpha.} Similar to MOBILE, we forbid the automatic entropy tuning trick in some datasets.

    \begin{table}[h]
        \centering
        \caption{Hyperparameters of DAMO used in the NeoRL datasets.}
        \resizebox{\linewidth}{!}{
        \begin{tabular}{lccccc}
            \toprule
             Task Name                  & $\alpha$ & $K$ & $l_{r}$ of actor & Offline data ratio & Auto-alpha
             \\ \midrule 
             Halfcheetah-L         & 0.3 & 5 & $1 \times 10^{-4}$ & 0.05 & True
             \\
             Hopper-L              & 1 & 5 & $1 \times 10^{-4}$ & 0.05 & True
             \\
             Walker2d-L            & 2.5 & 1 & $1 \times 10^{-4}$ & 0.1 & True
             \\ \midrule 
             Halfcheetah-M         & 0.5 & 3 & $1 \times 10^{-4}$ & 0.05 & True
             \\
             Hopper-M              & 3 & 5 & $5 \times 10^{-5}$ & 0.3 & False 
             \\
             Walker2d-M            & 3.5 & 1 & $1 \times 10^{-5}$ & 0.3 & True
             \\ \midrule 
             Halfcheetah-H         & 3 & 3 & $1 \times 10^{-4}$ & 0.05 & True
             \\
             Hopper-H              & 5 & 5 & $1 \times 10^{-5}$ & 0.05 & False 
             \\
             Walker2d-H            & 3 & 3 & $1 \times 10^{-4}$ & 0.3  & True
             \\ \bottomrule
        \end{tabular}
        }
        \label{tab-hyperparameter in neorl}
    \end{table}

    \subsection{Computing Infrastructure}
    \label{Appendix-Computing Infrastructure}
    The experimental setup includes a GeForce GTX 3090 GPU and an Intel(R) Xeon(R) Gold 6330 CPU at 2.00GHz. The implementation is built upon the OfflineRL-Kit library, with DAMO adopting the software libraries and frameworks specified therein. 

\end{document}